\documentclass{article}

\usepackage{arxiv}

\usepackage[utf8]{inputenc} 
\usepackage[T1]{fontenc}    
\usepackage{hyperref}       
\usepackage{url}            
\usepackage{booktabs}       
\usepackage{amsfonts}       
\usepackage{nicefrac}       
\usepackage{microtype}      
\usepackage{lipsum}

\usepackage{bm}
\usepackage{amsmath}
\usepackage{amssymb}
\usepackage{amsfonts}
\usepackage{graphicx}
\def\vec#1{\mathbf{#1}}
\usepackage{algorithm}
\usepackage{algorithmic}
\usepackage{comment}

\DeclareMathOperator*{\argmin}{arg\,min}

\title{Meta-learning to Calibrate Gaussian Processes with Deep Kernels for Regression Uncertainty Estimation}

\author{
  Tomoharu Iwata\\
  NTT Communication Science Laboratories, NTT Corporation\\
  \And
  Atsutoshi Kumagai\\
  NTT Computer and Data Science Laboratories, NTT Corporation\\
}
\date{}

\begin{document}
\maketitle

\begin{abstract}
Although Gaussian processes (GPs) with deep kernels have been succesfully used for meta-learning in regression tasks, its uncertainty estimation performance can be poor. We propose a meta-learning method for calibrating deep kernel GPs for improving regression uncertainty estimation performance with a limited number of training data. The proposed method meta-learns how to calibrate uncertainty using data from various tasks by minimizing the test expected calibration error, and uses the knowledge for unseen tasks. We design our model such that the adaptation and calibration for each task can be performed without iterative procedures, which enables effective meta-learning. In particular, a task-specific uncalibrated output distribution is modeled by a GP with a task-shared encoder network, and it is transformed to a calibrated one using a cumulative density function of a task-specific Gaussian mixture model (GMM). By integrating the GP and GMM into our neural network-based model, we can meta-learn model parameters in an end-to-end fashion. Our experiments demonstrate that the proposed method improves uncertainty estimation performance while keeping high regression performance compared with the existing methods using real-world datasets in few-shot settings.
\end{abstract}

\section{Introduction}

Estimating the uncertainty of a prediction in regression tasks
is important for machine learning systems.
When the confidence in a prediction is low,
decision-making can be passed on human experts to improve
reliability~\cite{kang2021statistical,begoli2019need,michelmore2020uncertainty}.
Accurate uncertainty estimation is also beneficial
for Bayesian optimization~\cite{shahriari2015taking},
reinforcement learning~\cite{yu2020mopo,malik2019calibrated,garcia2012safe,chua2018deep},
and active learning~\cite{gal2017deep}.
Many machine learning models, such as neural networks and Gaussian processes (GP)~\cite{rasmussen2005gaussian},
achieve high accuracy but have no uncertainty estimates
or poorly calibrated~\cite{guo2017calibration,tran2019calibrating,marx2022modular}.
To estimate uncertainty, several approaches have been proposed~\cite{abdar2021review,song2019distribution,sahoo2021reliable}.
For example, calibration methods transform the output
of a trained prediction model using a non-decreasing function
such that the predicted and empirical probabilities match.
However, the existing methods require many data for training.
In real-world applications,
enough data can be unavailable in tasks of interest,
and preparing sufficient data for each task requires high costs and is time-consuming.

\begin{figure}[t!]
  \centering
  \begin{tabular}{cc}
  \includegraphics[width=24em]{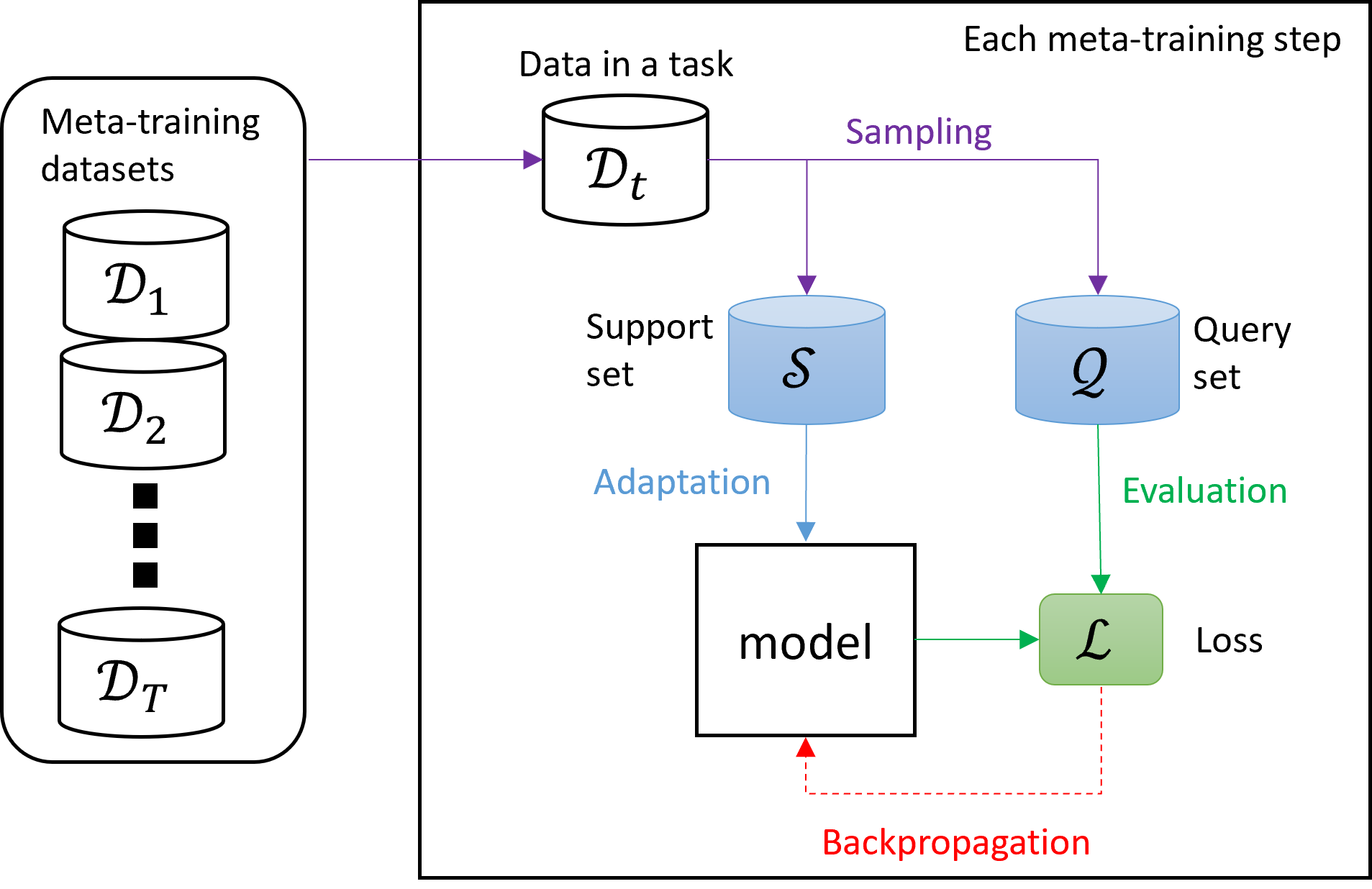}&
  \includegraphics[width=15em]{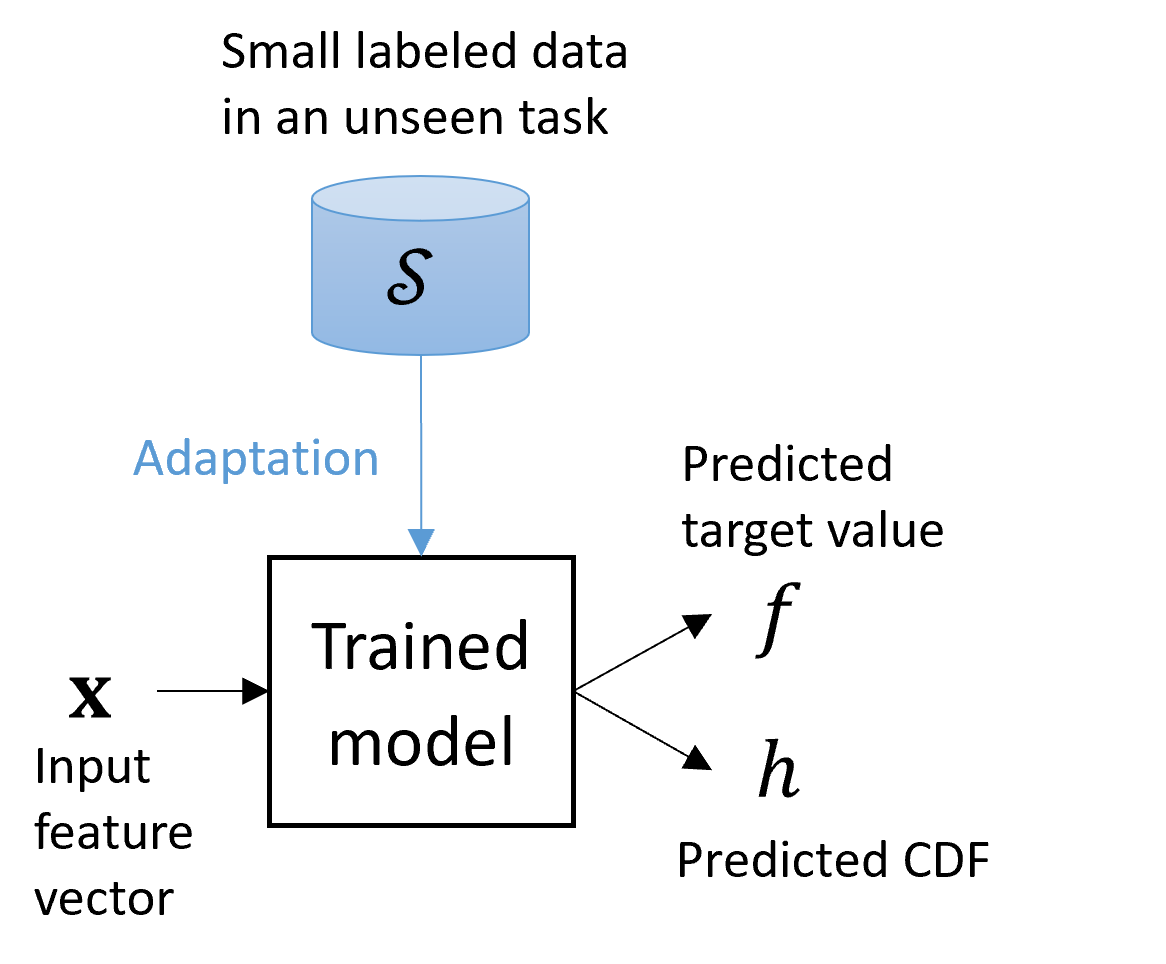}  \\
  (a) Meta-training phase & (b) Meta-test phase\\
  \end{tabular}
  \caption{Problem setting. (a) In the meta-training phase, given meta-training datasets from various tasks, our model is meta-learned. For each step, support and query sets are randomly sampled from a randomly sampled dataset. The support set is used for adapting the model to the task. The query set is used to evaluate the model for minimizing the test prediction and calibration losses based on backpropagation. (b) In the meta-test phase, given a small number of labeled data $\mathcal{S}$ in an unseen task, we predict target value $f$ and its CDF $h$ using the trained model adapted to $\mathcal{S}$.}
  \label{fig:framework}
\end{figure}

We propose a meta-learning method for estimating
uncertainty in regression tasks with a small amount of data.
Meta-learning methods
based on GPs with deep kernels~\cite{wilson2016deep}
have been successfully used for
improving the regression performance on unseen tasks
with few data by learning to learn from various tasks~\cite{tossou2019adaptive,fortuin2019meta,harrison2020meta,patacchiola2020bayesian,iwata2021few}.
However, these existing meta-learning methods are not designed
for uncertainty estimation.
The proposed method trains a regression model using data in multiple tasks
such that the expected uncertainty estimation performance
is improved when the model is adapted and calibrated to each task.
Figure~\ref{fig:framework} illustrates our problem setting.

Our model consists of task-shared and task-specific components.
With the task-shared components, we can share knowledge between tasks.
With the task-specific components, we can handle the heterogeneity of tasks.
Our meta-learning framework is formulated as a bilevel optimization,
where the inner optimization corresponds to the adaptation and calibration of the task-specific components,
and the outer optimization corresponds to the estimation of the task-shared components.
Since the inner optimization is repeated for each outer optimization epoch,
an effective inner optimization that allows backpropagation is essential.
Although the adaptation and calibration are usually performed by iterative procedures based on gradient descent~\cite{kuleshov2018accurate},
backpropagation through such procedures is costly in terms of memory,
and the total number of iterations must be kept small~\cite{finn2017model}.
On the other hand, we design our regression model such that the inner optimization is performed without iterative procedures.

In our model, an uncalibrated output distribution is modeled by a GP
with a task-shared encoder network.
With the GP, the task-adapted output distribution can be obtained in a closed form.
Since the uncertainty estimation performance of the output distribution can be low,
we calibrate it using a cumulative density function (CDF)
of a task-specific Gaussian mixture model (GMM), which is a non-decreasing and differentiable function,
such that the output distribution matches the empirical one.
The parameters of the GMM can be determined without iterative procedures.
In the outer optimization, the task-shared parameters are estimated by minimizing
the test expected calibration and regression errors
using a stochastic gradient descent method
with an episodic training framework~\cite{ravi2016optimization}.
Since the adaptation and calibration of our model are differentiable,
we can backpropagate the losses through them to update the task-shared components
in an end-to-end manner.


The main contributions of this paper are as follows:
1) We propose a meta-learning method for regression uncertainty estimation that directly minimizes the test expected calibration error.
2) We develop differentiable adaptation and calibration functions that do not require iterative optimization procedures.
3) We empirically demonstrate that the proposed method outperforms the existing uncertainty estimation methods.

\section{Related work}

Many methods for regression uncertainty estimation have been proposed.
Calibration methods map an output of a pretrained model to a calibrated one~\cite{platt1999probabilistic,kuleshov2018accurate,vovk2020conformal,niculescu2005predicting,chung2021uncertainty,marx2022modular}.
Quantile regression estimates quantiles of the output distribution, often by the pinball loss~\cite{fasiolo2021fast,salem2020prediction,tagasovska2019single,pearce2018high,jeon2016quantile}.
Some methods use models that output density function estimates~\cite{zhao2020individual,skafte2019reliable,lakshminarayanan2017simple,maddox2019simple,hernandez2015probabilistic,blundell2015weight,futami2021loss,koller2009probabilistic}.
We employed the calibration approach since
it has been used in a wide variety of applications,
and it enables us to design a model that can perform the adaptation and calibration without iterative procedures.
The existing calibration methods use
calibration models that require iterative optimization procedures~\cite{cui2020calibrated},
such as isotonic regression~\cite{kuleshov2018accurate},
or that are not differentiable~\cite{marx2022modular}.
In contrast, our calibration model is differentiable and obtained without iterative procedures,
and therefore, we can perform meta-learning effectively.

Most existing meta-learning methods are developed to improve
regression or classification
performance~\cite{schmidhuber:1987:srl,bengio1991learning,finn2017model,vinyals2016matching,snell2017prototypical},
but not uncertainty estimation performance.
There are some meta-learning methods for estimating classification uncertainty~\cite{bohdal2021meta,yang2022calibrating,kim2022calibration},
but not for regression uncertainty.
Neural processes~\cite{garnelo2018neural,garnelo2018conditional,kim2018attentive,nguyen2022transformer} are a meta-learning approach for few-shot regression with uncertainty estimation.
Since their prediction is based on fully parametric models,
they are less flexible in adapting to the given data~\cite{iwata2021few}.
Functional PAC-optimal hyper-posterior (F-PACOH)~\cite{rothfuss2021meta} is
a meta-learning method
to improve uncertainty estimation by alleviating meta-overfitting
in the case of few meta-training tasks
by regularizing priors in the function space.
F-PACOH is not a calibration method, and
it can be combined with the proposed method to directly minimize
the test expected calibration error.
Although \cite{bohdal2021meta} proposed a differentiable expected calibration error
for classification tasks,
it is inapplicable to regression tasks.
In meta-learning, models that can adapt to each task without iterative procedures
have been successfully used for classification and regression tasks using
linear or GP regression models~\cite{bertinetto2018meta,iwata2021few,patacchiola2020bayesian}.
However, they do not consider uncertainty calibration.
In addition, unlike these existing methods, the proposed method
achieves non-iterative task adaptation for a two-step method,
i.e., regression by GP
and its calibration by non-decreasing function,
by properly connecting differentiable approaches
for regression and calibration.

\section{Preliminaries: Calibration}

Let $h(y|\vec{x}):\mathcal{Y}\times\mathcal{X}\rightarrow [0,1]$
be the cumulative distribution function (CDF) of target value $y\in\mathcal{Y}\subseteq\mathbb{R}$
at input feature vector $\vec{x}\in\mathcal{X}$.
We can quantify the uncertainty of the prediction by the CDF,
which is the probability that the target value is less than $y$ given input $\vec{x}$.
The inverse function of the CDF $h^{-1}(p|\vec{x}):[0,1]\times\mathcal{X}\rightarrow\mathcal{Y}$
is used to denote the quantile function.
The CDF is well calibrated when it matches with the empirical CDF as sample size $N$ goes to
infinity~\cite{dawid1982well,kuleshov2018accurate}:
\begin{align}
  \frac{1}{N}\sum_{n=1}^{N}I(y_{n}\leq h^{-1}(p|\vec{x}_{n}))\rightarrow p,
\end{align}
as $N\rightarrow\infty$ for all $p\in[0,1]$.

Given data $\{(\vec{x}_{n},y_{n})\}_{n=1}^{N}$,
uncalibrated CDF $h_{\mathrm{U}}(y|\vec{x})$
is calibrated by non-decreasing function $r:[0,1]\rightarrow[0,1]$,
where $r(p_{(1)}),\dots,r(p_{(N)})$
are evenly spaced on the unit interval,
and $p_{(n)}=h_{\mathrm{U}}(y_{(n)}|\vec{x}_{(n)})$
is the $n$th smallest uncalibrated CDF in the given data.
An example of such a non-decreasing function is the following
empirical calibration model~\cite{marx2022modular}:
\begin{align}
  r_{\mathrm{emp}}(p)=\frac{n}{N}\quad \text{if $p\in[p_{(n)},p_{(n+1)})$},
    \label{eq:r_emp}
\end{align}
where $\frac{n}{N}$ is the empirical CDF between $p_{(n)}$ and $p_{(n+1)}$.
Figure~\ref{fig:r}(a) shows an example of $r_{\mathrm{emp}}$.
Although the empirical calibration model can calibrate uncertainty on given data,
its generalization performance for uncertainty estimation can be poor,
especially when only a small number of data are given.
A calibrated CDF is given by $h(y|\vec{x}) = r(h_{\mathrm{U}}(y|\vec{x}))$
using uncalibrated model $h_{\mathrm{U}}$ and calibration model $r$.

\section{Proposed method}

In Section~\ref{sec:problem}, we present our problem formulation
tackled in this paper.
In Section~\ref{sec:model},
we propose our model that outputs task-specific calibrated CDF $h$ given a small number of data.
In Section~\ref{sec:meta-learning}, we present meta-learning procedures for our model that improve
uncertainty estimation performance.

\subsection{Problem formulation}
\label{sec:problem}

In the meta-training phase,
we are given meta-training datasets $\{\mathcal{D}_{t}\}_{t=1}^{T}$ from $T$ tasks,
where $\mathcal{D}_{t}=\{(\vec{x}_{tn},y_{tn})\}_{n=1}^{N_{t}}$
is the dataset of the  $t$th task,
$\vec{x}_{tn}\in\mathcal{X}$ is the $n$th feature vector,
$y_{tn}\in\mathbb{R}$ is its target value,
and $N_{t}$ is the number of instances.
In the meta-test phase,
we are given a small number of labeled instances $\mathcal{S}=\{(\vec{x}^{\mathrm{S}}_{n},y^{\mathrm{S}}_{n})\}_{n=1}^{N^{\mathrm{S}}}$,
which is called the support set, from an unseen meta-test task that is different from
the meta-training tasks.
We are also given unlabeled instances, which are called the query set, from the meta-test task.
Our aim is to improve the test uncertainty estimation performance and
test regression performance on the query set in such meta-test tasks.

\subsection{Model}
\label{sec:model}

\begin{figure*}[t!]
  \centering
  \includegraphics[width=25em]{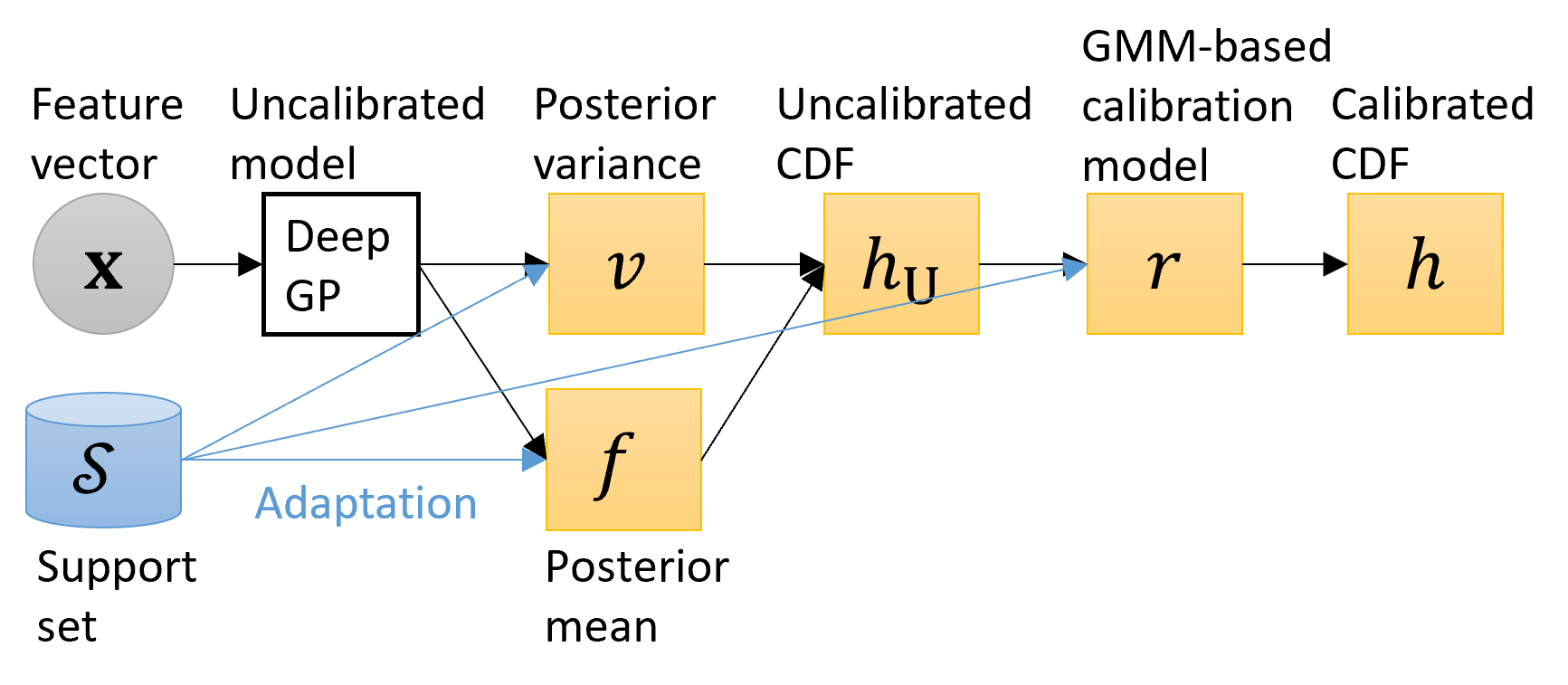}
  \caption{Proposed model. The task-shared/task-specific components in the proposed model are represented by white/yellow boxes, and the forwarding procedure is represented by black arrows. 
    Posterior mean $f$ and variance $v$ are modeled by GP with deep kernels, or deep GP.
    Uncalibrated CDF $h_{\mathrm{U}}$ is calculated from them.
    Calibrated CDF $h$ is obtained by transforming uncalibrated CDF $h_{\mathrm{U}}$ by GMM-based calibration model $r$. Posterior mean $f$, posterior variance $v$, and calibration model $r$ are adapted for each task using support set $\mathcal{S}$ as represented by blue arrows.}
  \label{fig:model}
\end{figure*}

We design our model to output predicted target value $f$ and calibrated CDF $h$
such that it can be adapted to support set $\mathcal{S}$ without iterative optimization.
An uncalibrated CDF is modeled by the GP with deep kernels as described in Section~\ref{sec:gp},
and its calibration is presented in Section~\ref{sec:calibration}.
Figure~\ref{fig:model} illustrates our model.

\subsubsection{Uncalibrated model}
\label{sec:gp}

We model an uncalibrated CDF using GP with
deep kernels~\cite{wilson2016deep},
$\mathrm{GP}(\mu(\vec{x}),k(\vec{x},\vec{x}'))$.
Here,
$\mu$ is the mean function based on a feed-forward neural network,
and $k$ is the following Gaussian kernel with an additive noise term,
\begin{align}
  k(\vec{x},\vec{x}')=\exp\left(-\frac{1}{2}\parallel g(\vec{x})-g(\vec{x}')\parallel^{2}\right)+\beta\delta(\vec{x},\vec{x}'),
  \label{eq:k}
\end{align}
where $g$ is an encoder based on a feed-forward neural network,
$\beta>0$ is an observation's additive noise parameter,
and $\delta(\vec{x},\vec{x}')$ is the Kronecker delta function,
$\delta(\vec{x},\vec{x}')=1$ if $\vec{x}$ and $\vec{x}'$ are identical, and zero otherwise.
Neural networks, $\mu$ and $g$, and kernel parameter $\beta$
are shared across tasks, by which
we can accumulate knowledge that is useful for uncertainty estimation in different tasks in model parameters, and use it for unseen tasks.
With neural network $g$, we can flexibly encode input features
such that the target value is accurately predicted by the
GP in the encoded space.
With the GP,
the posterior probability density distribution of the target value
adapted to support set $\mathcal{S}$ is calculated
in a closed form by the following Gaussian,
\begin{align}
  p_{\mathrm{U}}(y|\vec{x};\mathcal{S}) = \mathcal{N}(y|f(\vec{x};\mathcal{S}),v(\vec{x};\mathcal{S})),
  \label{eq:q_u}
\end{align}
where
\begin{align}
  f(\vec{x};\mathcal{S})=\mu(\vec{x})+\vec{k}^{\top}\vec{K}^{-1}(\vec{y}^{\mathrm{S}}-\vec{m}), \quad
  v(\vec{x};\mathcal{S}) = k(\vec{x},\vec{x})-\vec{k}^{\top}\vec{K}^{-1}\vec{k},
  \label{eq:f}  
\end{align}
are its mean and variance,
$\vec{y}^{\mathrm{S}}=(y_{n}^{\mathrm{S}})_{n=1}^{N^{\mathrm{S}}}\in\mathbb{R}^{N^{\mathrm{S}}}$
is the target vector of the support instances,
$\vec{K}\in\mathbb{R}^{N^{\mathrm{S}}\times N^{\mathrm{S}}}$ is the matrix of the kernel function evaluated among
the support instances,
$\vec{K}_{nn'}=k(\vec{x}^{\mathrm{S}}_{n},\vec{x}^{\mathrm{S}}_{n'})$,
$\vec{k}=(k(\vec{x},\vec{x}^{\mathrm{S}}_{n}))_{n=1}^{N^{\mathrm{S}}}\in\mathbb{R}^{N^{\mathrm{S}}}$
is the vector of the kernel function evaluated between $\vec{x}$ and the support instances,
$\vec{m}=(\mu(\vec{x}^{\mathrm{S}}_{n}))_{n=1}^{N^{\mathrm{S}}}\in\mathbb{R}^{N^{\mathrm{S}}}$
is the vector of the mean function evaluated on the support instances,
and $\mathcal{N}(\cdot|\mu,\sigma^{2})$
is the Gaussian probability distribution function with mean $\mu$ and variance $\sigma^{2}$.
Posterior mean $f(\vec{x};\mathcal{S})$ corresponds to the predicted target value.
Using Eq.~(\ref{eq:q_u}), the uncalibrated CDF is given by
\begin{align}
  h_{\mathrm{U}}(y|\vec{x};\mathcal{S})
    =\int_{-\infty}^{y}p_{\mathrm{U}}(y'|\vec{x};\mathcal{S})dy'
  =\frac{1}{2}\left(1+\mathrm{erf}\left(\frac{y-f(\vec{x};\mathcal{S})}{\sqrt{2v(\vec{x};\mathcal{S})}}\right)\right),
  \label{eq:h_u}
\end{align}
which can be calculated using error function $\mathrm{erf}(y)= \frac{2}{\sqrt{\pi}}\int_{0}^{y}\exp(-y'^{2})dy$.
The gradient of the error function, which is required for backpropagation,
follows immediately from its definition $\frac{\partial\mathrm{erf}(y)}{\partial y}=\frac{2}{\sqrt{\pi}}\exp(-y^{2})$,
which is given in a closed form.

\subsubsection{Calibration model}
\label{sec:calibration}

\begin{figure}[t!]
  \centering
      {\tabcolsep=1em\begin{tabular}{cc}
    \includegraphics[width=15em]{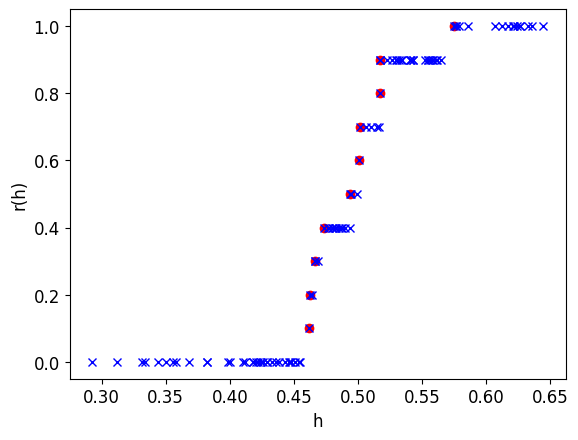}&        
    \includegraphics[width=15em]{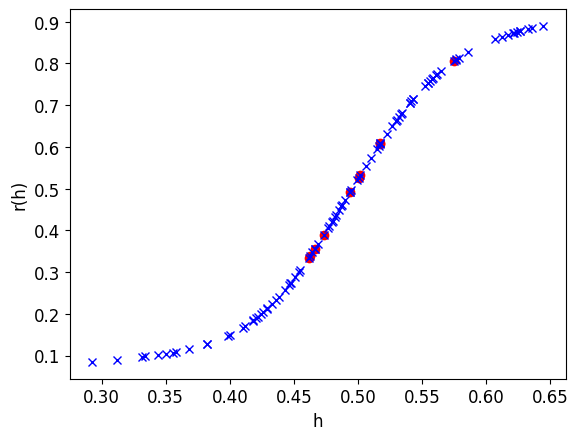}\\
    (a) Empirical calibration model &    
    (b) Our calibration model \\
    \includegraphics[width=15em]{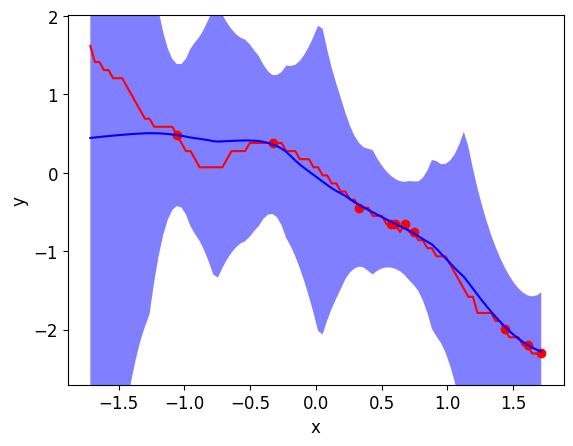}&
    \includegraphics[width=15em]{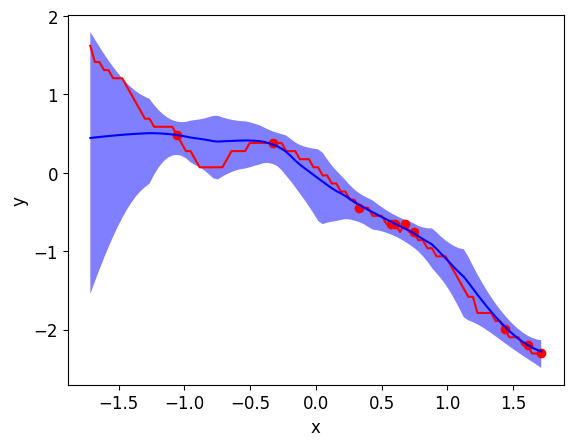}\\
    (c) Uncalibrated model &
    (d) Calibrated model\\
  \end{tabular}}
      \caption{
        (a) Empirical calibration model.
        The horizontal axis is the uncalibrated CDF, and the vertical axis is its calibrated one with the calibration model $r_{\mathrm{emp}}$ in Eq.~(\ref{eq:r_emp}).
        A red point shows a support instance, and a blue point shows a query instance. Here, we used the AHM dataset (see Section~\ref{sec:experiments} for details, and we also used the same data in (b,c) and Figure~\ref{fig:cdf}).
        (b) Our calibration model $r$ in Eq.~(\ref{eq:r}).
        (c) Our uncalibrated model adapted to a support set, and (d) its calibrated one with our calibration model. The horizontal axis is the input feature value, and the vertical axis is the target value. A red point shows a support instance, a red line shows true target values, a blue line shows predicted target values, and a blue area shows estimated 95\% confidence intervals.
        Here, we used the AHM dataset (see Section~\ref{sec:experiments} for details, and we also used the same data in (b,c) and Figure~\ref{fig:cdf}).
}
  \label{fig:r}
\end{figure}

We calibrate uncalibrated CDF $h_{\mathrm{U}}$ using a task-specific non-decreasing function,
for which we use the CDF of the following Gaussian mixture model,
\begin{align}
  q(h';\mathcal{S})
  =\frac{1}{N^{\mathrm{s}}} \sum_{(\vec{x}^{\mathrm{S}},y^{\mathrm{S}})\in\mathcal{S}}\mathcal{N}(h'|h_{\mathrm{U}}(y^{\mathrm{S}}|\vec{x}^{\mathrm{S}};\mathcal{S}),\sigma^{2}),
  \label{eq:gmm}
\end{align}
where
the number of components is the number of support instances $N^{\mathrm{S}}$,
the mean of each component is the uncalibrated CDF in Eq.~(\ref{eq:h_u}) at each support instance, 
variance $\sigma^{2}$ is a task-shared parameter to be trained,
and the mixture weight is $\frac{1}{N^{\mathrm{S}}}$.
Its CDF is given by
\begin{align}
  r(h';\mathcal{S})=\int_{-\infty}^{h'}q(h'';\mathcal{S})dh''
  =\frac{1}{2N^{\mathrm{S}}}\sum_{(\vec{x}^{\mathrm{S}},y^{\mathrm{S}})\in\mathcal{S}}\left(1+\mathrm{erf}\left(\frac{h'-h_{\mathrm{U}}(y^{\mathrm{S}}|\vec{x}^{\mathrm{S}};\mathcal{S})}{\sqrt{2}\sigma}\right)\right),  
  \label{eq:r}
\end{align}
where it is a non-decreasing function.
As variance $\sigma^{2}$ goes to zero,
our calibration model in Eq.~(\ref{eq:r}) gets close to the empirical calibration model
in Eq.~(\ref{eq:r_emp}).
Although the empirical calibration model is not differentiable,
our calibration model is differentiable.
By using the Gaussian mixture model,
we can obtain a differentiable non-decreasing function for calibration without iterative optimization.
An example of calibration model $r$ is shown in Figure~\ref{fig:r}(b).
It is smooth and non-decreasing, and its slope is steep where the support instances exist.
Although uncalibrated model $h_{\mathrm{U}}(y|\vec{x})$ in Eq.~(\ref{eq:h_u}) can model only a Gaussian distribution,
calibrated model $r(h(y|\vec{x}))$ using Eq.~(\ref{eq:r}) can model a non-Gaussian distribution.

The existing calibration methods usually prepare data for calibration 
that are different from data for training an uncalibrated prediction model to avoid overfitting~\cite{kuleshov2018accurate}.
However, since we consider a situation where the number of support instances is small,
splitting them into two can be unprofitable.
Therefore, we use support set $\mathcal{S}$ for adapting both uncalibrated and calibration models.
Since we train task-shared parameters such that the test calibration loss is minimized as described in the next subsection,
we can alleviate the overfitting of the calibration.
To additionally reduce the risk of overfitting,
we use the following mixture of calibrated and uncalibrated CDFs,
\begin{align}
  h(y|\vec{x},\mathcal{S})
    =\alpha h_{\mathrm{U}}(y|\vec{x};\mathcal{S})
    +
  (1-\alpha)r(h_{\mathrm{U}}(y|\vec{x};\mathcal{S});\mathcal{S}),
  \label{eq:h}
\end{align}
where $\alpha\in[0,1]$ is a task-shared weight parameter
that can be trained automatically using meta-training datasets.
When $r(p)$ is a non-increasing function, $\alpha p+(1-\alpha)r(p)$ is also a non-increasing function.

Figures~\ref{fig:r}(c,d) show the predicted target values
and their uncertainty with our uncalibrated and calibrated models.
In both models, the uncertainty was small,
and the predicted target values were close to true ones
around the support instances, 
and the uncertainty was large where the support instances were located far away.
These results are reasonable. 
However, the uncertainty by the uncalibrated model was generally large, and poorly calibrated.
By the calibration, the uncertainty was tightened while most of the true target values were within the 95\% confidence interval,
and well-calibrated.
Since the uncalibrated model can model only a Gaussian distribution,
the confidence interval was always symmetrical around the predicted target values.
On the other hand, the calibrated model can represent an unsymmetric confidence interval.

\begin{figure}[t!]
  \centering
  {\tabcolsep=1em\begin{tabular}{cc}  
    \includegraphics[width=15em]{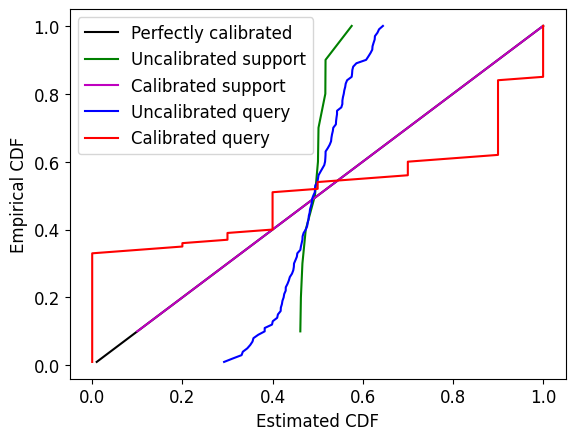}
    &
    \includegraphics[width=15em]{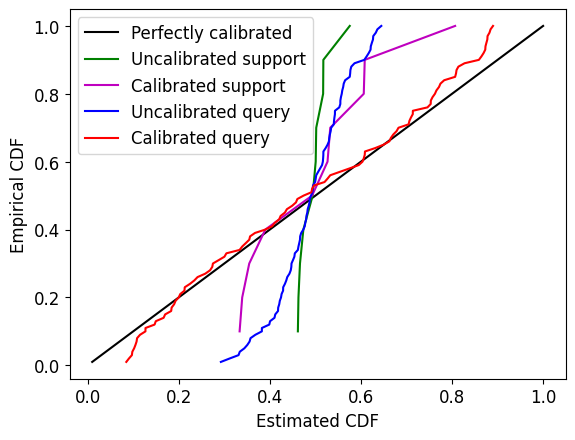}\\    
    (a) Empirical calibration model
    &
  (b) Our calibration model\\
  \end{tabular}}
  \caption{Estimated CDFs of uncalibrated/calibrated models and empirical CDFs on support/query instances by empirical calibration model $r_{\rm{emp}}$ in Eq.~(\ref{eq:r_emp}) (a) and by our calibration model (b). The black line shows the perfectly calibrated estimated CDF and the empirical CDF. The green/purple line shows the estimated uncalibrated CDF and the empirical CDF on support/query instances.
    The blue/red line shows the estimated calibrated CDF and the empirical CDF on support/query instances.}
  \label{fig:cdf}
\end{figure}

Figure~\ref{fig:cdf} shows the relationship between the estimated and empirical CDFs when calibrated with
empirical calibration model $r_{\mathrm{emp}}$ in Eq.~(\ref{eq:r_emp}) and with our calibration model.
When the estimated and empirical CDFs match as shown in the black line,
it is perfectly calibrated.
The calibration of the empirical CDF on the support instances is easily achievable
via the empirical calibration model as shown in (a).
However, its estimated CDF on the query instances was poorly calibrated due to overfitting,
which is undesirable since our aim is to improve the test uncertainty estimation performance.
On the other hand, the estimated CDF on the query instances by our calibrated model
was closer to the empirical CDF than that by the uncalibrated model
and that by the empirical calibration model.

\subsection{Meta-learning}
\label{sec:meta-learning}

The parameters $\bm{\phi}$ shared across tasks in our model
are 
parameters in encoder network $g$ in Eq.~(\ref{eq:k}), parameters in neural network-based mean function $\mu$ in Eq.~(\ref{eq:f}),
kernel parameter $\beta$ in Eq.~(\ref{eq:k}),
variance $\sigma^{2}$ in the calibration model in Eq.~(\ref{eq:gmm}),
and
weight $\alpha$ in the calibrated CDF in Eq.~(\ref{eq:h}).
We train $\bm{\phi}$ such that
the test target prediction performance
and test uncertainty estimation performance are improved.
Let $\mathcal{Q}$ be the query set, which contains held-out data different from support set $\mathcal{S}$ in a task.
For evaluating the test target prediction performance of prediction model $f$ on the meta-training tasks,
we use the following regression loss,
$\mathcal{L}_{\mathrm{R}}(f,\mathcal{Q})=\frac{1}{N^{\mathrm{Q}}}\sum_{(\vec{x},y)\in\mathcal{Q}}\parallel f(\vec{x})-y\parallel^{2}$,
which is the mean squared error between true and predicted target values
over the query instances.
For evaluating the test uncertainty estimation performance of CDF $h$,
we use the following calibration loss,
\begin{align}
  \mathcal{L}_{\mathrm{C}}(h,\mathcal{Q})=\frac{1}{N^{\mathrm{Q}}}\sum_{n=1}^{N^{\mathrm{Q}}}\left| p_{(n)}-\frac{n}{N^{\mathrm{Q}}}\right|,
  \label{eq:L_C}
\end{align}
where $p_{(n)}$ is the $n$th smallest CDF estimated by $h$ in the query instances,
and $\frac{n}{N^{\mathrm{Q}}}$ is the empirical CDF of the query instances at $p_{(n)}$.
When the calibration loss is small,
the empirical and estimated CDFs are close.
For the total loss, we use the following
weighted average of the regression and calibration losses,
\begin{align}
  \mathcal{L}(f,h,\mathcal{Q})=\lambda\mathcal{L}_{\mathrm{R}}(h,\mathcal{Q})
  +(1-\lambda)\mathcal{L}_{\mathrm{C}}(f,\mathcal{Q}),
  \label{eq:L}
\end{align}
where $0\leq\lambda\leq 1$ is a hyperparameter that can be determined
depending on its application.
In our experiments, the performance was not sensitive to $\lambda$
unless $\lambda$ was almost zero or one.
Although proper scoring rules~\cite{gneiting2007strictly},
such as the negative log-likelihood and pinball loss~\cite{jeon2016quantile},
can consider both regression and calibration performance
in a measurement indirectly, they fail to minimize calibration error
because of overfitting~\cite{kumar2018trainable}.
Therefore, we directly use the regression and calibration errors
for the objective function in Eq.~(\ref{eq:L})
as in~\cite{kumar2018trainable}.

We estimate parameters $\bm{\phi}$ by minimizing the following
expected total loss
calculated using meta-training datasets $\{\mathcal{D}_{t}\}_{t=1}^{T}$,
\begin{align}
  \hat{\bm{\phi}}=\argmin_{\phi}\mathbb{E}_{t}\mathbb{E}_{\mathcal{S},\mathcal{Q}\sim\mathcal{D}_{t}}\left[
    \mathcal{L}(f(\cdot;\mathcal{S}),h(\cdot;\mathcal{S}),\mathcal{Q})\right],
  \label{eq:phi_hat}
\end{align}
where $\mathbb{E}_{t}$ is the expectation over meta-training tasks,
$\mathbb{E}_{\mathcal{S},\mathcal{Q}\sim\mathcal{D}_{t}}$ is the expectation over support and query sets that are randomly generated without overlap
from a meta-training dataset $\mathcal{D}_{t}$,
$f(\cdot;\mathcal{S})$
is the target prediction model adapted to support set $\mathcal{S}$ in Eq.~(\ref{eq:f}),
and
$h(\cdot;\mathcal{S})$
is the calibrated CDF adapted to support set $\mathcal{S}$ in Eq.~(\ref{eq:h}).
We can improve the test performance by Eq.~(\ref{eq:phi_hat})
since the support set for adaptation and calibration
is different from the query set for evaluation.
When the task distribution in meta-training data
is the same as that in meta-test data,
Eq.~(\ref{eq:phi_hat}) improves
the generalization performance for the meta-test data.
Since both models $f$ and $h$ are differentiable,
we can minimize it using a stochastic gradient descent method~\cite{kingma2014adam}.
Algorithm~\ref{alg:train} shows the meta-learning procedures of our model.
The expectation in Eq.~(\ref{eq:phi_hat}) is approximated by the Monte Carlo method
by randomly sampling tasks, support, and query sets from the meta-training datasets in Lines 2--4.
The time complexity cubically increases with the number of support instances $N^{\mathrm{S}}$
due to the inverse in Eqs.~(\ref{eq:f}).
Since we consider situations where the number of support instances is small,
our model can be optimized efficiently.

\begin{algorithm}[t!]
  \centering
  \caption{Meta-learning procedures of our model.}
  \label{alg:train}
  \begin{algorithmic}[1]
    \renewcommand{\algorithmicrequire}{\textbf{Input:}}
    \renewcommand{\algorithmicensure}{\textbf{Output:}}
    \REQUIRE{Meta-training data $\{\mathcal{D}\}_{t=1}^{T}$,
      support set size $N^{\mathrm{S}}$,
      query set size $N^{\mathrm{Q}}$.}
    \ENSURE{Trained task-shared parameters $\bm{\phi}$.}
    \WHILE{End condition is not satisfied}
    \STATE Randomly select task index $t$ from $\{1,\cdots,T\}$.    
    \STATE Randomly sample $N^{\mathrm{S}}$ instances from $\mathcal{D}_{t}$ for support set $\mathcal{S}$.
    \STATE Randomly sample $N^{\mathrm{Q}}$ instances from $\mathcal{D}_{t}\setminus\mathcal{S}$ for query set $\mathcal{Q}$.
    \STATE Calculate uncalibrated PDF $p_{\mathrm{U}}$ adapted to support set $\mathcal{S}$ by Eq.~(\ref{eq:q_u}).
    \STATE Calculate uncalibrated CDF $h_{\mathrm{U}}$ by Eq.~(\ref{eq:h_u}).
    \STATE Calculate calibration model $r$ adapted to support set $\mathcal{S}$ by Eq.~(\ref{eq:r}).
    \STATE Calculate calibrated CDF $h$ by Eq.~(\ref{eq:h}).
    \STATE Calculate loss $\mathcal{L}$ on query set $\mathcal{Q}$ by Eq.~(\ref{eq:L}).
    \STATE Update model parameters $\bm{\phi}$ using the gradient of the loss by a stochastic gradient method.
    \ENDWHILE
  \end{algorithmic}
\end{algorithm}

\section{Experiments}
\label{sec:experiments}

\subsection{Data}

To evaluate the proposed method,
we used the following five data sets:
AHM, MAT, School, Tafeng, and Sales.
AHM and MAT are the spatial data of the annual heat-moisture index
and mean annual temperature in North America.
School is a benchmark dataset in multi-task regression~\cite{bakker2003task,argyriou2006multi},
which contains the examination scores of students from different schools (tasks).
Sales and Tafeng contain weekly purchased quantities of products~\cite{he2019efficient}.

AHM and MAT were the spatial data of the annual heat-moisture index
and mean annual temperature in North America,
which were obtained from \url{https://sites.ualberta.ca/~ahamann/data/climatena.html}.
We used 365 non-overlapping regions (tasks), where the size of each region was $100 \times 100$km,
and the target values were observed at $1 \times 1$km grid squares in each region.
The number of features in AHM was one (longitude),
and the number of features in MAT was three (longitude, latitude, and elevation).
School is a benchmark dataset in multi-task regression~\cite{bakker2003task,argyriou2006multi},
which contains the examination scores of students from 139 schools (tasks).
We used schools with 60 or more students, ending up with 109 tasks.
The number of features was 28, and the average number of instances per task was 129.
Sales contains weekly purchased quantities of 811 products over 52 weeks~\cite{tan2014time}.
We obtained the dataset from the UCI repository.
Following~\cite{he2019efficient},
we used the sales of five previous weeks for each product as features,
and the sales for the current week (task) as targets.
The number of features was five, the number of tasks was 47,
and the number of instances per task was 811.
Tafeng is another grocery shopping dataset that contains transactions of 23,812 products
over four months.
We build the data in a similar fashion to Sales.
The number of features was five, the number of tasks was 13,
and the number of instances per task was 23,812.

For each dataset, we randomly split the tasks,
where 60\% of them were used for meta-training,
20\% for meta-validation,
and the remaining for meta-test.

\subsection{Compared methods}

We compared the proposed method (Ours) with the following methods:
meta-learning with deep kernel learning (MDKL)~\cite{iwata2021few},
neural process (NP)~\cite{garnelo2018conditional},
simultaneous quantile regression (SQR)~\cite{tagasovska2019single},
neural network-based Gaussian model (NG),
Bayesian dropout (BD)~\cite{gal2016dropout},
Gaussian process (GP),
and meta-learning versions of SQR, NG, and BD (MSQR, MNG, and MBD)
based on model-agnostic meta-learning (MAML)~\cite{finn2017model}.
SQR, NG, BD, and GP are single-task methods,
which train models for each task.
MDKL, NP, MSQR, MNG, MBD, and Ours are meta-learning methods,
which train task-shared neural networks using meta-training datasets.

With MDKL, the GP with deep kernels is used,
where the neural network and kernel parameters are trained
by maximizing the expected test likelihood on query sets.
NP is a neural network-based meta-learning method, where
a task representation is encoded using the support set,
and the predictive mean and variance are calculated using the task representation.
With MDKL, the GP with deep kernels is used.
In NP and MDKL,
the model parameters are trained
by maximizing the expected test likelihood on query sets.
SQR is a neural network model that outputs a quantile target value given
a feature vector and quantile level,
which is trained by minimizing the pinball loss
for various quantile levels $p\in[0.1,0.2,\dots,0.9]$.
NG is a neural network model that outputs the predictive mean and variance
of a Gaussian distribution of a target value given a feature vector,
which is trained by maximizing the likelihood.
With BD, a neural network regression model is trained by minimizing
the mean squared error with dropout,
and the mean and variance of a target value
are estimated by sampling multiple target values with dropout.
The dropout rate was $0.3$, and the number of samples was $30$.
With GP, the RBF kernel was used, where the kernel parameters were trained by maximizing
the expected test likelihood using meta-training datasets.
With MAML-based methods,
the task adaptation was performed
by the gradient descent method with learning rate $10^{-2}$ and five epochs.

\subsection{Settings}

For encoder network $g$ in the proposed method, MDKL, and NP,
we used a three-layered feed-forward neural network with 32 hidden and output units.
For mean function $f$ in the proposed method and MDKL, and for neural networks in comparing methods,
we used four-layered feed-forward neural networks with 32 hidden units.
We optimized the models using Adam~\cite{kingma2014adam} with learning rate $10^{-2}$,
and a batch size of 32 tasks.
The number of meta-training epochs was 1,000, and
the meta-validation data were used for early stopping.
We implemented the methods with PyTorch~\cite{paszke2019pytorch}.
The number of support instances per task was $\{10,20,30\}$,
and the number of query instances was $30$.

\subsection{Measurements}

For the evaluation measurement of the target prediction and
uncertainty estimation performance,
we used the mean squared error (MSE)
and expected calibration error (ECE)~\cite{naeini2015obtaining,guo2017calibration}
on query sets in the meta-test tasks.
The MSE is calculated by the squared error
between true and predicted targets,
$\mathrm{MSE}=\frac{1}{N^{\mathrm{Q}}}\sum_{(\vec{x},y)\in\mathcal{Q}}\parallel y-f(\vec{x})\parallel^{2}$.
The ECE is a widely-used measurement for uncertainty estimation performance,
which is calculated by the absolute error
between quantile level $p$ and
empirical probability $\hat{p}$ that the target value is below the quantile function,
$\mathrm{ECE}=\frac{1}{|\mathcal{P}|}\sum_{p\in\mathcal{P}}|p-\hat{p}(p,h,\mathcal{Q})|$,
where
$\hat{p}(p,h,\mathcal{Q})=\frac{1}{N^{\mathrm{Q}}}\sum_{(\vec{x},y)\in\mathcal{Q}}I(y\leq h^{-1}(p|\vec{x}))$
and $\mathcal{P}=\{0.1,0.2,\dots,0.9\}$.
For evaluating both performances simultaneously,
we used the total error (TE), which is the average of the MSE and ECE, 
$\mathrm{TE}=\frac{\mathrm{MSE}+{\mathrm{ECE}}}{2}$.
We averaged the TE, MSE, and ECE over ten experiments with different splits of meta-training,
validation, and test data.

\subsection{Results}

\begin{table*}[t!]
  \centering
  \caption{Test total error, which is the average of MSE and ECE, with different numbers of support instances $N^{\mathrm{S}}$. Values in bold typeface are not statistically significantly different at the 5\% level from the best performing method in each row according to a paired t-test.}
  \label{tab:total_error}
  \begin{normalsize}
    {\tabcolsep=0.5em
      \begin{tabular}{lrrrrrrrrrrr}  
  \hline
  Data & $N^{\mathrm{S}}$ & Ours & MDKL & NP & MSQR & MNG & MBD & SQR & NG & BD & GP \\
  \hline
AHM & 10 & {\bf 0.167} & 0.184 & 0.262 & 0.809 & 0.857 & 0.870 & 0.532 & 0.564 & 0.248 & 0.202\\
AHM & 20 & {\bf 0.097} & 0.120 & 0.229 & 0.796 & 0.850 & 0.858 & 0.517 & 0.556 & 0.195 & 0.127\\
AHM & 30 & {\bf 0.077} & 0.104 & 0.214 & 0.789 & 0.840 & 0.846 & 0.504 & 0.541 & 0.186 & 0.107\\
\hline
MAT & 10 & {\bf 0.123} & 0.133 & {\bf 0.124} & 0.272 & 0.382 & 0.325 & 0.540 & 0.431 & 0.222 & 0.134\\
MAT & 20 & {\bf 0.094} & 0.106 & 0.114 & 0.271 & 0.383 & 0.325 & 0.526 & 0.415 & 0.171 & 0.107\\
MAT & 30 & {\bf 0.077} & 0.095 & 0.112 & 0.276 & 0.380 & 0.328 & 0.525 & 0.410 & 0.155 & 0.096\\
  \hline
School & 10 & {\bf 0.383} & {\bf 0.381} & {\bf 0.381} & 0.392 & 0.415 & 0.478 & 0.662 & 0.581 & 0.756 & {\bf 0.382}\\
School & 20 & {\bf 0.380} & {\bf 0.379} & {\bf 0.377} & 0.383 & 0.410 & 0.477 & 0.645 & 0.552 & 0.683 & {\bf 0.379}\\
School & 30 & {\bf 0.373} & {\bf 0.371} & {\bf 0.375} & 0.377 & 0.404 & 0.466 & 0.632 & 0.533 & 0.616 & {\bf 0.373}\\
  \hline
Tafeng & 10 & {\bf 0.957} & {\bf 0.928} & {\bf 1.036} & {\bf 0.910} & {\bf 1.129} & {\bf 0.984} & {\bf 1.190} & {\bf 1.302} & {\bf 1.115} & {\bf 0.910}\\
Tafeng & 20 & {\bf 0.882} & {\bf 0.994} & {\bf 1.091} & {\bf 0.802} & 1.051 & {\bf 0.976} & {\bf 1.156} & {\bf 1.249} & {\bf 1.140} & {\bf 0.967}\\
Tafeng & 30 & {\bf 0.134} & 0.186 & 0.196 & {\bf 0.141} & 0.250 & 0.238 & 0.202 & {\bf 0.284} & 0.338 & 0.216\\
  \hline
Sales & 10 & {\bf 0.081} & 0.100 & {\bf 0.082} & {\bf 0.086} & 0.275 & 0.124 & 0.411 & 0.381 & 0.195 & 0.102\\
Sales & 20 & {\bf 0.081} & 0.094 & {\bf 0.082} & {\bf 0.085} & 0.267 & 0.124 & 0.397 & 0.340 & 0.153 & 0.097\\
Sales & 30 & {\bf 0.082} & 0.093 & {\bf 0.083} & {\bf 0.087} & 0.272 & 0.121 & 0.395 & 0.331 & 0.148 & 0.098\\
  \hline
\end{tabular}}
  \end{normalsize}
  \end{table*}

\begin{table*}[t!]
  \centering
  \caption{Test mean squared error (MSE) with different numbers of support instances $N^{\mathrm{S}}$. Values in bold typeface are not statistically significantly different at the 5\% level from the best performing method in each row according to a paired t-test.}
  \label{tab:mse}
  \begin{normalsize}
    {\tabcolsep=0.5em
\begin{tabular}{lrrrrrrrrrrr}  
  \hline
  Data & $N^{\mathrm{S}}$ & Ours & MDKL & NP & MSQR & MNG & MBD & SQR & NG & BD & GP \\
  \hline
AHM & 10 & {\bf 0.223} & {\bf 0.229} & 0.399 & 1.413 & 1.500 & 1.407 & 0.792 & 0.962 & 0.360 & 0.265\\
AHM & 20 & {\bf 0.104} & {\bf 0.101} & 0.340 & 1.391 & 1.488 & 1.383 & 0.767 & 0.952 & 0.264 & 0.115\\
AHM & 30 & 0.069 & {\bf 0.064} & 0.315 & 1.378 & 1.470 & 1.363 & 0.745 & 0.924 & 0.250 & 0.069\\
\hline
MAT & 10 & {\bf 0.140} & {\bf 0.142} & 0.162 & 0.454 & 0.654 & 0.467 & 0.831 & 0.738 & 0.315 & 0.146\\
MAT & 20 & {\bf 0.097} & {\bf 0.095} & 0.143 & 0.452 & 0.656 & 0.466 & 0.809 & 0.715 & 0.236 & 0.099\\
MAT & 30 & {\bf 0.073} & {\bf 0.073} & 0.139 & 0.463 & 0.653 & 0.472 & 0.807 & 0.710 & 0.211 & 0.079\\
\hline
School & 10 & {\bf 0.706} & {\bf 0.702} & {\bf 0.702} & 0.723 & 0.770 & 0.753 & 1.072 & 1.079 & 1.307 & {\bf 0.703}\\
School & 20 & {\bf 0.700} & {\bf 0.697} & {\bf 0.695} & 0.707 & 0.760 & 0.753 & 1.038 & 1.030 & 1.174 & {\bf 0.698}\\
School & 30 & {\bf 0.688} & {\bf 0.683} & 0.692 & 0.699 & 0.752 & 0.729 & 1.014 & 0.999 & 1.044 & {\bf 0.690}\\
\hline
Tafeng & 10 & {\bf 1.819} & {\bf 1.672} & {\bf 1.909} & {\bf 1.707} & {\bf 2.012} & {\bf 1.788} & {\bf 2.178} & {\bf 2.365} & {\bf 2.040} & {\bf 1.635}\\
Tafeng & 20 & {\bf 1.667} & {\bf 1.803} & {\bf 2.011} & {\bf 1.489} & {\bf 1.861} & {\bf 1.771} & {\bf 2.128} & {\bf 2.258} & {\bf 2.090} & {\bf 1.747}\\
Tafeng & 30 & 0.187 & 0.189 & {\bf 0.209} & {\bf 0.168} & 0.266 & {\bf 0.299} & {\bf 0.223} & {\bf 0.327} & {\bf 0.490} & {\bf 0.250}\\
\hline
Sales & 10 & {\bf 0.089} & {\bf 0.090} & {\bf 0.089} & 0.091 & 0.338 & 0.130 & 0.557 & 0.572 & 0.258 & 0.098\\
Sales & 20 & 0.086 & 0.085 & {\bf 0.081} & 0.085 & 0.322 & 0.128 & 0.530 & 0.489 & 0.189 & 0.093\\
Sales & 30 & {\bf 0.087} & {\bf 0.087} & {\bf 0.085} & {\bf 0.088} & 0.330 & 0.121 & 0.529 & 0.474 & 0.180 & 0.095\\
\hline
    \end{tabular}}
    \end{normalsize}
  \end{table*}

\begin{table*}[t!]
  \centering
  \caption{Test expected calibration error (ECE) with different numbers of support instances $N^{\mathrm{S}}$. Values in bold typeface are not statistically significantly different at the 5\% level from the best performing method in each row according to a paired t-test.}
  \label{tab:ece}
  \begin{normalsize}
    {\tabcolsep=0.5em  
\begin{tabular}{lrrrrrrrrrrr}  
  \hline
  Data & $N^{\mathrm{S}}$ & Ours & MDKL & NP & MSQR & MNG & MBD & SQR & NG & BD & GP \\
  \hline
AHM & 10 & {\bf 0.111} & 0.139 & 0.125 & 0.205 & 0.213 & 0.333 & 0.271 & 0.166 & 0.136 & 0.140\\
AHM & 20 & {\bf 0.090} & 0.140 & 0.118 & 0.201 & 0.211 & 0.332 & 0.268 & 0.161 & 0.125 & 0.139\\
AHM & 30 & {\bf 0.086} & 0.144 & 0.113 & 0.201 & 0.209 & 0.330 & 0.262 & 0.158 & 0.122 & 0.145\\
  \hline
MAT & 10 & 0.106 & 0.123 & {\bf 0.086} & {\bf 0.089} & 0.109 & 0.183 & 0.250 & 0.124 & 0.130 & 0.122\\
MAT & 20 & 0.090 & 0.117 & {\bf 0.084} & 0.089 & 0.110 & 0.183 & 0.244 & 0.115 & 0.105 & 0.115\\
MAT & 30 & {\bf 0.082} & 0.116 & {\bf 0.084} & 0.090 & 0.108 & 0.183 & 0.244 & 0.111 & 0.098 & 0.114\\
  \hline
School & 10 & {\bf 0.060} & {\bf 0.061} & {\bf 0.060} & {\bf 0.061} & {\bf 0.060} & 0.204 & 0.251 & 0.083 & 0.205 & {\bf 0.060}\\
School & 20 & {\bf 0.060} & {\bf 0.061} & {\bf 0.060} & {\bf 0.059} & {\bf 0.060} & 0.202 & 0.252 & 0.073 & 0.191 & {\bf 0.059}\\
School & 30 & 0.059 & 0.060 & {\bf 0.057} & {\bf 0.055} & {\bf 0.056} & 0.203 & 0.251 & 0.067 & 0.187 & {\bf 0.056}\\
  \hline
Tafeng & 10 & {\bf 0.094} & 0.184 & 0.164 & {\bf 0.113} & 0.247 & 0.180 & 0.201 & 0.239 & 0.190 & 0.185\\
Tafeng & 20 & {\bf 0.097} & 0.185 & 0.170 & 0.114 & 0.242 & 0.181 & 0.185 & 0.239 & 0.191 & 0.187\\
Tafeng & 30 & {\bf 0.081} & 0.183 & 0.184 & 0.113 & 0.233 & 0.178 & 0.180 & 0.247 & 0.187 & 0.183\\
  \hline
Sales & 10 & {\bf 0.073} & 0.110 & {\bf 0.075} & {\bf 0.081} & 0.213 & 0.118 & 0.266 & 0.190 & 0.132 & 0.107\\
Sales & 20 & {\bf 0.075} & 0.103 & 0.084 & {\bf 0.085} & 0.212 & 0.120 & 0.264 & 0.190 & 0.116 & 0.102\\
Sales & 30 & {\bf 0.076} & 0.099 & {\bf 0.080} & {\bf 0.087} & 0.213 & 0.120 & 0.261 & 0.188 & 0.116 & 0.102\\
\hline
    \end{tabular}}
    \end{normalsize}
  \end{table*}

\begin{figure*}[t!]
  \centering
  {\tabcolsep=0em\begin{tabular}{ccc}
  \includegraphics[width=15em]{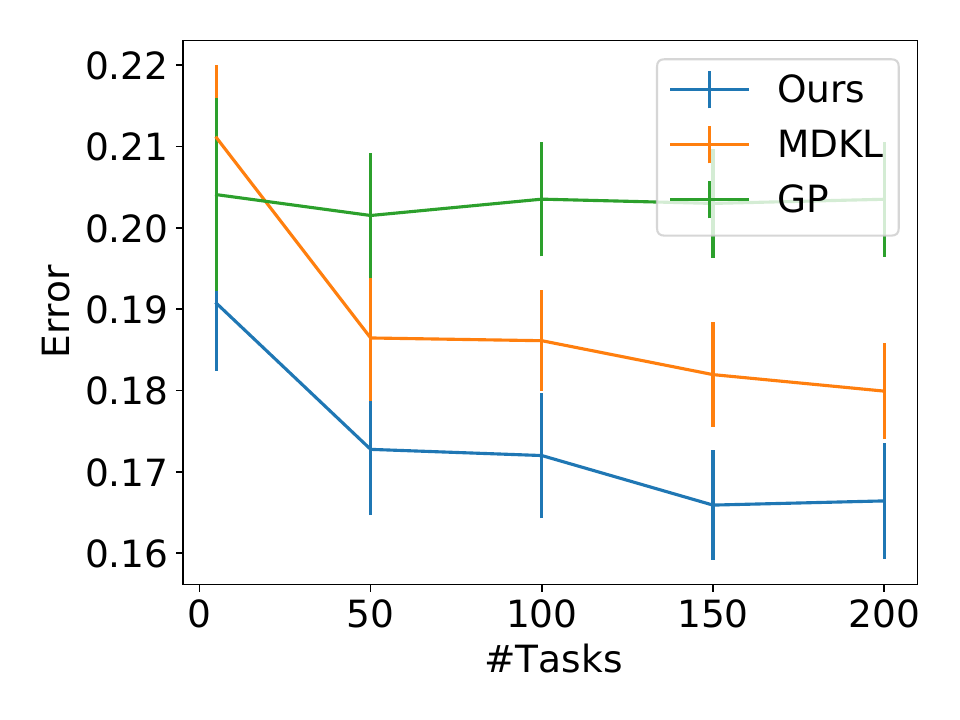}&
  \includegraphics[width=15em]{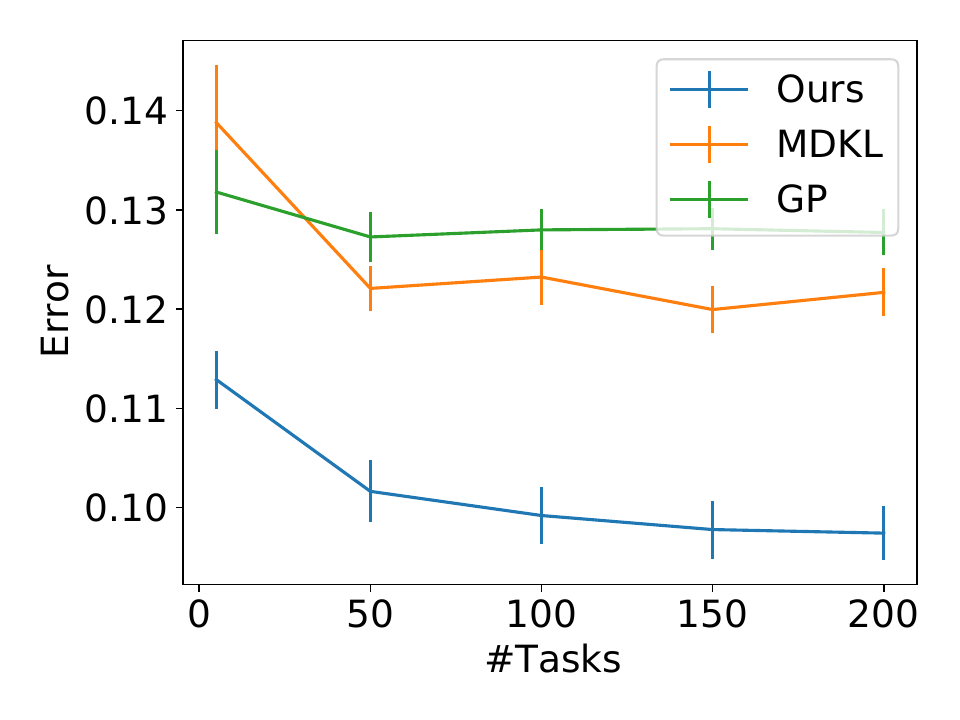}&
  \includegraphics[width=15em]{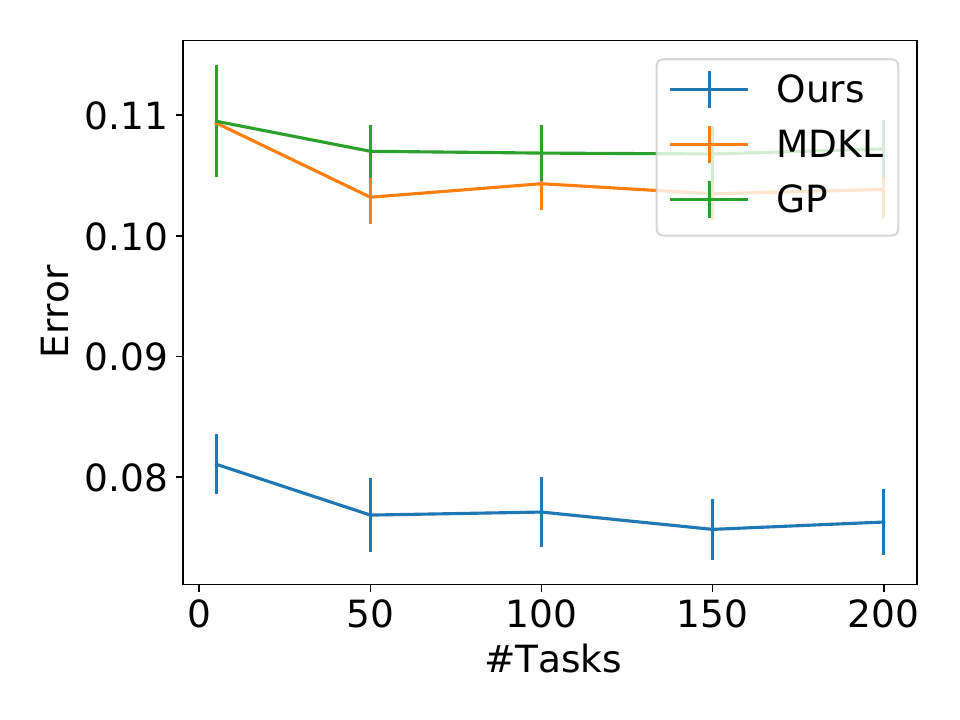}\\
  (a) $N^{\mathrm{S}}=10$ & (b) $N^{\mathrm{S}}=20$ & (c) $N^{\mathrm{S}}=30$ \\ 
  \end{tabular}}
  \caption{Test total error with different numbers of training tasks with the AHM datasets. The bar shows the standard error.}
  \label{fig:err_task}
\end{figure*}

\begin{table*}[t!]
  \centering
  \caption{Ablation study of the proposed method. The test total errors are shown. `w/o-Net' is without neural networks in the deep kernel GP,
    `w/o-$r$' is without calibration model $r$,
    `w/o-$\mathcal{L}_{\mathrm{C}}$' is without calibration loss $\mathcal{L}_{\mathrm{C}}$ in the objective function in Eq.~(\ref{eq:L}),
    `w/o-Mix' is without the mixing of the uncalibrated CDF in Eq.~(\ref{eq:h}) or fixing $\alpha=1$,
    `w/-Split' is with support set splitting for adaptation and calibration,
    and 'w/-$r_{\mathrm{emp}}$ uses the empirical calibration model in Eq.~(\ref{eq:r_emp}) for calibration. Values in bold typeface are not statistically significantly different at the 5\% level from the best performing method in each row according to a paired t-test.}
  \label{tab:ablation}
  \begin{normalsize}
    {\tabcolsep=0.5em
\begin{tabular}{lrrrrrrrrrr}  
  \hline
  Data & $N^{\mathrm{S}}$ & Ours & w/o-Net & w/o-$r$ & w/o-$\mathcal{L}_{\mathrm{C}}$ & w/o-Mix & w/-Split & w/-$r_{\mathrm{emp}}$\\
  \hline
AHM & 10 & {\bf 0.167} & 0.185 & 0.176 & 0.195 & {\bf 0.167} & 0.255 & 0.172\\
AHM & 20 & {\bf 0.097} & 0.103 & 0.115 & 0.138 & {\bf 0.099} & 0.160 & 0.102\\
AHM & 30 & {\bf 0.077} & 0.078 & 0.094 & 0.122 & {\bf 0.077} & 0.116 & 0.083\\
\hline
MAT & 10 & {\bf 0.123} & 0.136 & 0.133 & 0.149 & 0.126 & 0.165 & 0.128\\
MAT & 20 & {\bf 0.094} & 0.099 & 0.104 & 0.119 & {\bf 0.093} & 0.117 & 0.098\\
MAT & 30 & {\bf 0.077} & 0.083 & 0.093 & 0.105 & 0.080 & 0.099 & 0.084\\
\hline
School & 10 & {\bf 0.383} & 0.531 & {\bf 0.384} & 0.464 & 0.399 & {\bf 0.381} & 0.385\\
School & 20 & {\bf 0.380} & 0.506 & {\bf 0.379} & 0.460 & 0.390 & {\bf 0.377} & 0.391\\
School & 30 & {\bf 0.373} & 0.470 & {\bf 0.373} & 0.453 & 0.379 & {\bf 0.374} & 0.383\\
\hline
Tafeng & 10 & {\bf 0.957} & {\bf 1.159} & {\bf 0.882} & {\bf 0.961} & {\bf 0.875} & {\bf 0.939} & {\bf 0.772}\\
Tafeng & 20 & {\bf 0.882} & {\bf 1.094} & {\bf 0.517} & {\bf 0.939} & {\bf 0.889} & {\bf 0.897} & {\bf 0.832}\\
Tafeng & 30 & {\bf 0.134} & {\bf 0.146} & 0.159 & 0.215 & {\bf 0.155} & {\bf 0.155} & {\bf 0.146}\\
\hline
Sales & 10 & {\bf 0.081} & 0.144 & 0.086 & 0.161 & 0.087 & {\bf 0.081} & 0.089\\
Sales & 20 & {\bf 0.081} & 0.104 & 0.091 & 0.159 & 0.088 & {\bf 0.079} & 0.086\\
Sales & 30 & {\bf 0.082} & 0.093 & 0.085 & 0.159 & 0.085 & {\bf 0.083} & 0.086\\
\hline
    \end{tabular}}
    \end{normalsize}
\end{table*}

\begin{figure*}[t!]
  \centering
  {\tabcolsep=0em\begin{tabular}{ccc}
  \includegraphics[width=15em]{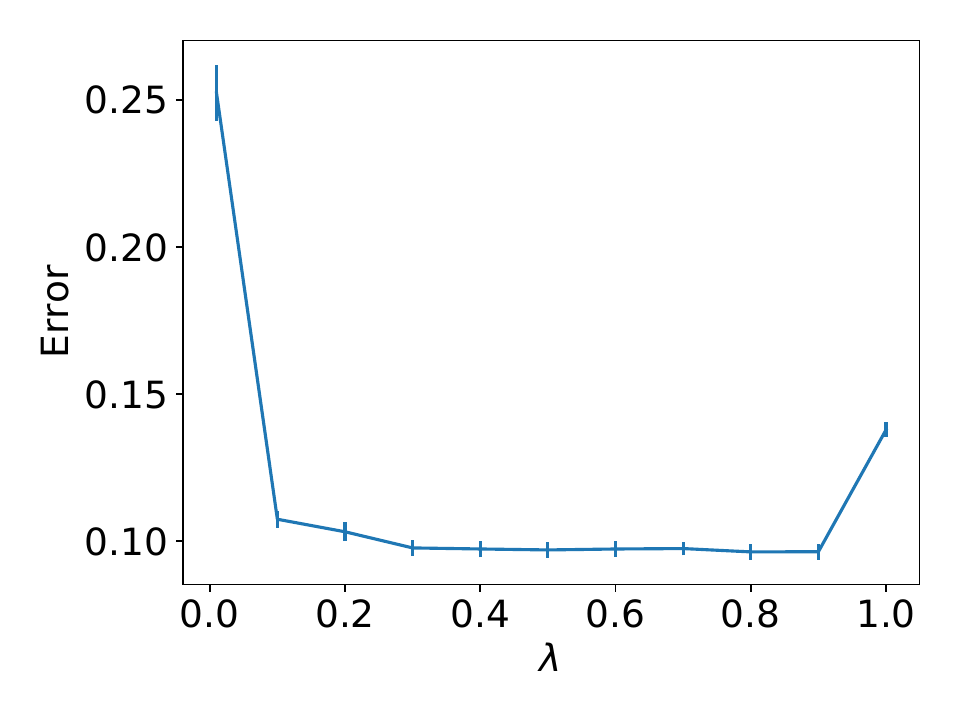}&
  \includegraphics[width=15em]{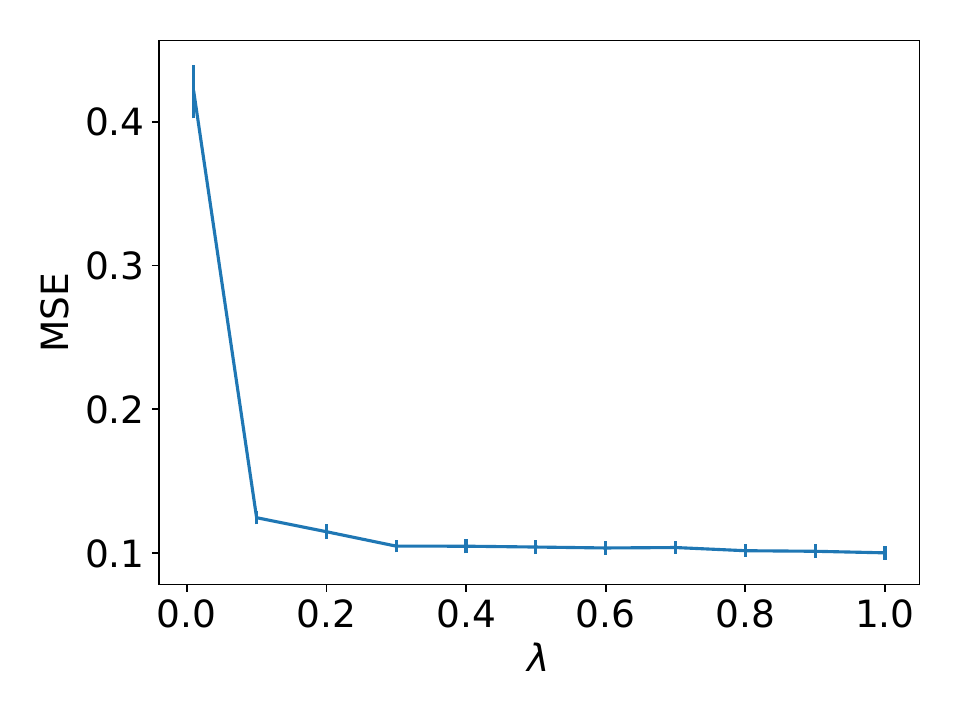}&
  \includegraphics[width=15em]{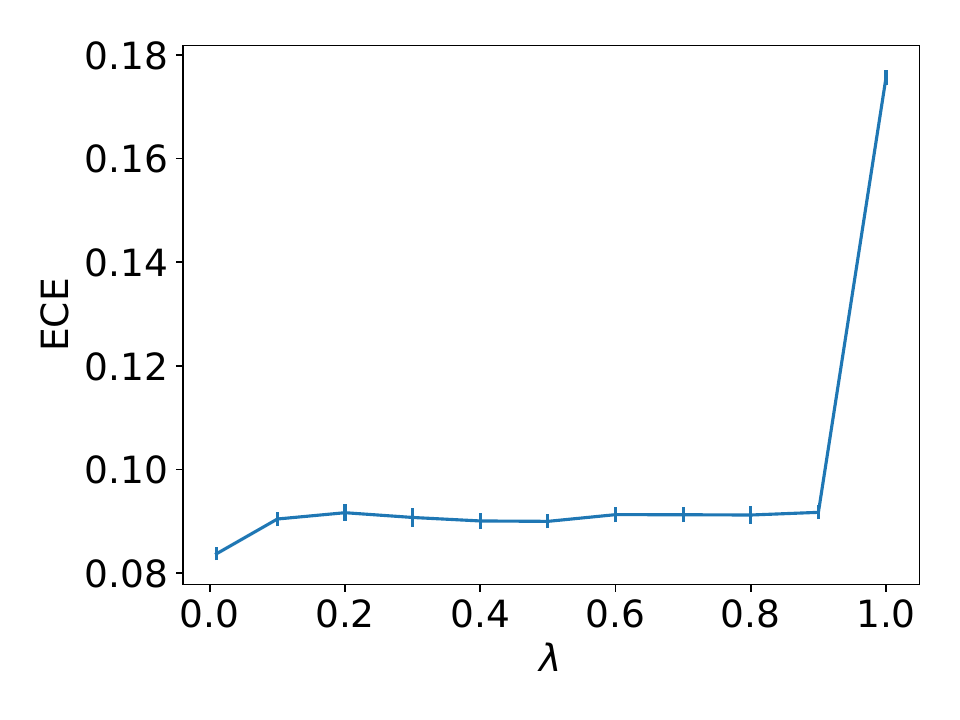}\\
  (a) Total error & (b) MSE & (c) ECE \\
  \end{tabular}}
  \caption{Test total error (a), MSE (b), and ECE (c) with different hyperparameters $\lambda$ with the AHM datasets. The bar shows the standard error.}
  \label{fig:err_lambda}
\end{figure*}

Tables~\ref{tab:total_error}, \ref{tab:mse} and \ref{tab:ece}
show the test total error, MSE and ECE.
The proposed method achieved the lowest test total error in all cases.
As the number of support instances increased, the performance generally rose.
MDKL corresponds to the proposed method without uncertainty calibration that minimizes the regression loss.
Although the MSEs by the proposed method were close to those by MDKL,
the ECEs by the proposed method were always better than those by MDKL.
This result demonstrates the effectiveness of our calibration model and calibration loss.
MAML-based meta-learning methods often improved the performance
compared with their single-task versions.
However, they failed to meta-learn well on the AHM dataset.
Since MAML uses a small number of gradient descent steps for adaptation,
the adaptation can be insufficient.
In contrast, since the proposed method performs task adaptation and calibration
without iterative procedures, its meta-learning becomes stable, resulting in its low errors.
The MSEs by the proposed method and MDKL were generally lower than the other methods.
This result indicates the efficacy of the GP-based closed-form solver for task adaptation in regression.

Figure~\ref{fig:err_task} shows that the test total error decreased
as the number of training tasks rose with the proposed method.
This result indicates that the proposed method
can meta-learn from various tasks, and use the knowledge for unseen tasks.
Table~\ref{tab:ablation} shows the ablation study result.
The better performance of the proposed method exhibits the effectiveness of each component.
Since the neural networks are used for sharing knowledge across tasks,
the proposed method without neural networks (w/o-Net) deteriorated the performance.
The proposed method without calibration models (w/o-$r$) corresponds to MDKL that minimizes
the expected test total error, which includes the expected test calibration error.
The increased error of w/o-$r$ shows
the effectiveness of our calibration model for calibrating the output distribution.
With the calibration loss, the expected test uncertainty estimation performance is directly optimized.
Therefore, removing the calibration loss (w/o-$\mathcal{L}_{\mathrm{C}}$)
resulted in an increase of the total error.
The results without the mixing of the uncalibrated CDF (w/o-Mix)
and those with the support set splitting (w/-Split)
indicate that mixing and unsplitting sometimes improve the performance.
Since the empirical calibration model (w/-$r_{\mathrm{emp}}$) is not differentiable, its performance was low.
Figure~\ref{fig:err_lambda} shows the test errors with different
$\lambda\in\{10^{-2},0.1,0.2,\dots,1.0\}$.
The test error was not sensitive to $\lambda$
unless $\lambda$ was close to zero or one.

We also compared with SQR, NG, BD, and GP methods that were trained
using meta-training datasets: ASAR, ANG, ABD, and AGP.
In AGP, sparse GPs~\cite{snelson2005sparse} were used
with task-shared pseudo-inputs and task-specific support instances.
Table~\ref{tab:total_error_all} shows their test total error.
When appropriate regression models are different across tasks,
these methods did not perform well.
The proposed method achieved the lowest test total error in all cases.

\begin{table*}[t!]
  \centering
  \caption{Test total error, which is the average of MSE and ECE, with different numbers of support instances $N^{\mathrm{S}}$. Values in bold typeface are not statistically significantly different at the 5\% level from the best performing method in each row according to a paired t-test.}
  \label{tab:total_error_all}
\begin{normalsize}
\begin{tabular}{lrrrrrr}
  \hline
  Data & $N^{\mathrm{S}}$ & Ours & ASQR & ANG & ABD & AGP\\
  \hline
AHM & 10 & {\bf 0.167} & 0.834 & 0.885 & 0.890 & 0.278\\
AHM & 20 & {\bf 0.097} & 0.824 & 0.876 & 0.883 & 0.189\\
AHM & 30 & {\bf 0.077} & 0.817 & 0.862 & 0.873 & 0.158\\
\hline
MAT & 10 & {\bf 0.123} & 0.272 & 0.385 & 0.325 & 0.131\\
MAT & 20 & {\bf 0.094} & 0.271 & 0.385 & 0.326 & 0.115\\
MAT & 30 & {\bf 0.077} & 0.277 & 0.386 & 0.328 & 0.106\\
\hline
School & 10 & {\bf 0.383} & 0.390 & 0.417 & 0.474 & 0.429\\
School & 20 & {\bf 0.380} & {\bf 0.386} & 0.409 & 0.468 & 0.425\\
School & 30 & {\bf 0.373} & 0.381 & 0.404 & 0.465 & 0.421\\
\hline
Tafeng & 10 & {\bf 0.957} & {\bf 0.911} & {\bf 1.081} & {\bf 0.769} & {\bf 1.136}\\
Tafeng & 20 & {\bf 0.882} & {\bf 0.828} & 1.060 & {\bf 0.975} & {\bf 1.123}\\
Tafeng & 30 & {\bf 0.134} & {\bf 0.145} & 0.266 & 0.208 & 0.190\\
\hline
Sales & 10 & {\bf 0.081} & {\bf 0.084} & 0.282 & 0.121 & 0.133\\
Sales & 20 & {\bf 0.081} & {\bf 0.082} & 0.275 & 0.110 & 0.133\\
Sales & 30 & {\bf 0.082} & {\bf 0.085} & 0.278 & 0.121 & 0.136\\
  \hline
\end{tabular}
\end{normalsize}
  \end{table*}

Table~\ref{tab:time} shows the computational time in seconds for meta-learning
using computers with Xeon Gold 6130 2.10GHz CPU, and 256GB memory.
Table~\ref{tab:time_test} shows the time in seconds
to output predicted target values and CDFs that are
adapted and calibrated to the support set.
Although the proposed method requires a long meta-learning time,
its inference time is short.

\begin{table}[t!]
  \centering
  \caption{Computational time in seconds for meta-learning on the AHM dataset with 20 support instances.}
  \label{tab:time}
\begin{normalsize}  
\begin{tabular}{rrrrrr}
  \hline  
  Ours & MDKL & NP & MSQR & MNG & MBD \\
  \hline
  588.745 & 264.324 & 83.448 & 4310.218 & 312.226 & 548.077 \\  
  \hline
\end{tabular}
\end{normalsize}  
\end{table}

\begin{table}[t!]
  \centering
  \caption{Computational time in seconds for inference on the AHM dataset with 20 support instances.}
  \label{tab:time_test}
  \begin{normalsize}  
\begin{tabular}{rrrrrrrrrr}
  \hline
Ours & MDKL & NP & MSQR & MNG & MBD & SQR & NG & BD & GP \\
  \hline
0.083 & 0.034 & 0.015 & 0.316 & 0.047 & 0.112 & 5.853 & 5.677 & 51.407 & 0.020\\
  \hline
\end{tabular}
\end{normalsize}  
\end{table}

\section{Conclusion}

We proposed a meta-learning method for regression uncertainty estimation,
and confirmed that the proposed method achieves better performance than the existing methods.
For future work, we plan to consider individual calibration instead of average calibration~\cite{chung2021beyond}
to approximate the conditional quantiles.
Also, we would like to extend the proposed method for classification tasks.

\bibliographystyle{abbrv}
\bibliography{neurocomp}

\begin{thebibliography}{10}

\bibitem{abdar2021review}
M.~Abdar, F.~Pourpanah, S.~Hussain, D.~Rezazadegan, L.~Liu, M.~Ghavamzadeh,
  P.~Fieguth, X.~Cao, A.~Khosravi, U.~R. Acharya, et~al.
\newblock A review of uncertainty quantification in deep learning: Techniques,
  applications and challenges.
\newblock {\em Information Fusion}, 76:243--297, 2021.

\bibitem{argyriou2006multi}
A.~Argyriou, T.~Evgeniou, and M.~Pontil.
\newblock Multi-task feature learning.
\newblock {\em Advances in Neural Information Processing Systems}, 2006.

\bibitem{bakker2003task}
B.~Bakker and T.~Heskes.
\newblock Task clustering and gating for {B}ayesian multitask learning.
\newblock {\em Journal of Machine Learning Research}, 4:83--99, 2003.

\bibitem{begoli2019need}
E.~Begoli, T.~Bhattacharya, and D.~Kusnezov.
\newblock The need for uncertainty quantification in machine-assisted medical
  decision making.
\newblock {\em Nature Machine Intelligence}, 1(1):20--23, 2019.

\bibitem{bengio1991learning}
Y.~Bengio, S.~Bengio, and J.~Cloutier.
\newblock Learning a synaptic learning rule.
\newblock In {\em International Joint Conference on Neural Networks}, 1991.

\bibitem{bertinetto2018meta}
L.~Bertinetto, J.~F. Henriques, P.~Torr, and A.~Vedaldi.
\newblock Meta-learning with differentiable closed-form solvers.
\newblock In {\em International Conference on Learning Representations}, 2018.

\bibitem{blundell2015weight}
C.~Blundell, J.~Cornebise, K.~Kavukcuoglu, and D.~Wierstra.
\newblock Weight uncertainty in neural network.
\newblock In {\em International Conference on Machine Learning}, pages
  1613--1622, 2015.

\bibitem{bohdal2021meta}
O.~Bohdal, Y.~Yang, and T.~Hospedales.
\newblock Meta-calibration: Meta-learning of model calibration using
  differentiable expected calibration error.
\newblock {\em arXiv preprint arXiv:2106.09613}, 2021.

\bibitem{chua2018deep}
K.~Chua, R.~Calandra, R.~McAllister, and S.~Levine.
\newblock Deep reinforcement learning in a handful of trials using
  probabilistic dynamics models.
\newblock {\em Advances in Neural Information Processing Systems}, 31, 2018.

\bibitem{chung2021uncertainty}
Y.~Chung, I.~Char, H.~Guo, J.~Schneider, and W.~Neiswanger.
\newblock Uncertainty toolbox: an open-source library for assessing,
  visualizing, and improving uncertainty quantification.
\newblock {\em arXiv preprint arXiv:2109.10254}, 2021.

\bibitem{chung2021beyond}
Y.~Chung, W.~Neiswanger, I.~Char, and J.~Schneider.
\newblock Beyond pinball loss: Quantile methods for calibrated uncertainty
  quantification.
\newblock {\em Advances in Neural Information Processing Systems},
  34:10971--10984, 2021.

\bibitem{cui2020calibrated}
P.~Cui, W.~Hu, and J.~Zhu.
\newblock Calibrated reliable regression using maximum mean discrepancy.
\newblock {\em Advances in Neural Information Processing Systems},
  33:17164--17175, 2020.

\bibitem{dawid1982well}
A.~P. Dawid.
\newblock The well-calibrated {B}ayesian.
\newblock {\em Journal of the American Statistical Association},
  77(379):605--610, 1982.

\bibitem{fasiolo2021fast}
M.~Fasiolo, S.~N. Wood, M.~Zaffran, R.~Nedellec, and Y.~Goude.
\newblock Fast calibrated additive quantile regression.
\newblock {\em Journal of the American Statistical Association},
  116(535):1402--1412, 2021.

\bibitem{finn2017model}
C.~Finn, P.~Abbeel, and S.~Levine.
\newblock Model-agnostic meta-learning for fast adaptation of deep networks.
\newblock In {\em International conference on machine learning}, pages
  1126--1135, 2017.

\bibitem{fortuin2019meta}
V.~Fortuin, H.~Strathmann, and G.~R{\"a}tsch.
\newblock Meta-learning mean functions for {G}aussian processes.
\newblock {\em arXiv preprint arXiv:1901.08098}, 2019.

\bibitem{futami2021loss}
F.~Futami, T.~Iwata, I.~Sato, M.~Sugiyama, et~al.
\newblock Loss function based second-order jensen inequality and its
  application to particle variational inference.
\newblock {\em Advances in Neural Information Processing Systems},
  34:6803--6815, 2021.

\bibitem{gal2016dropout}
Y.~Gal and Z.~Ghahramani.
\newblock Dropout as a {Bayesian} approximation: Representing model uncertainty
  in deep learning.
\newblock In {\em International Conference on Machine Learning}, pages
  1050--1059, 2016.

\bibitem{gal2017deep}
Y.~Gal, R.~Islam, and Z.~Ghahramani.
\newblock Deep {B}ayesian active learning with image data.
\newblock In {\em International Conference on Machine Learning}, pages
  1183--1192, 2017.

\bibitem{garcia2012safe}
J.~Garcia and F.~Fern{\'a}ndez.
\newblock Safe exploration of state and action spaces in reinforcement
  learning.
\newblock {\em Journal of Artificial Intelligence Research}, 45:515--564, 2012.

\bibitem{garnelo2018conditional}
M.~Garnelo, D.~Rosenbaum, C.~Maddison, T.~Ramalho, D.~Saxton, M.~Shanahan,
  Y.~W. Teh, D.~Rezende, and S.~A. Eslami.
\newblock Conditional neural processes.
\newblock In {\em International Conference on Machine Learning}, pages
  1704--1713, 2018.

\bibitem{garnelo2018neural}
M.~Garnelo, J.~Schwarz, D.~Rosenbaum, F.~Viola, D.~J. Rezende, S.~Eslami, and
  Y.~W. Teh.
\newblock Neural processes.
\newblock {\em arXiv preprint arXiv:1807.01622}, 2018.

\bibitem{gneiting2007strictly}
T.~Gneiting and A.~E. Raftery.
\newblock Strictly proper scoring rules, prediction, and estimation.
\newblock {\em Journal of the American statistical Association},
  102(477):359--378, 2007.

\bibitem{guo2017calibration}
C.~Guo, G.~Pleiss, Y.~Sun, and K.~Q. Weinberger.
\newblock On calibration of modern neural networks.
\newblock In {\em International Conference on Machine Learning}, pages
  1321--1330, 2017.

\bibitem{harrison2020meta}
J.~Harrison, A.~Sharma, and M.~Pavone.
\newblock Meta-learning priors for efficient online {B}ayesian regression.
\newblock In {\em Algorithmic Foundations of Robotics XIII: Proceedings of the
  13th Workshop on the Algorithmic Foundations of Robotics 13}, pages 318--337.
  Springer, 2020.

\bibitem{he2019efficient}
X.~He, F.~Alesiani, and A.~Shaker.
\newblock Efficient and scalable multi-task regression on massive number of
  tasks.
\newblock In {\em Proceedings of the AAAI Conference on Artificial
  Intelligence}, volume~33, pages 3763--3770, 2019.

\bibitem{hernandez2015probabilistic}
J.~M. Hern{\'a}ndez-Lobato and R.~Adams.
\newblock Probabilistic backpropagation for scalable learning of {B}ayesian
  neural networks.
\newblock In {\em International Conference on Machine Learning}, pages
  1861--1869, 2015.

\bibitem{iwata2021few}
T.~Iwata and Y.~Tanaka.
\newblock Few-shot learning for spatial regression via neural embedding-based
  {G}aussian processes.
\newblock {\em Machine Learning}, 111:1239--1257, 2022.

\bibitem{jeon2016quantile}
S.~Jeon, C.~J. Paciorek, and M.~F. Wehner.
\newblock Quantile-based bias correction and uncertainty quantification of
  extreme event attribution statements.
\newblock {\em Weather and Climate Extremes}, 12:24--32, 2016.

\bibitem{kang2021statistical}
D.~Y. Kang, P.~N. DeYoung, J.~Tantiongloc, T.~P. Coleman, and R.~L. Owens.
\newblock Statistical uncertainty quantification to augment clinical decision
  support: a first implementation in sleep medicine.
\newblock {\em NPJ Digital Medicine}, 4(1):1--9, 2021.

\bibitem{kim2018attentive}
H.~Kim, A.~Mnih, J.~Schwarz, M.~Garnelo, A.~Eslami, D.~Rosenbaum, O.~Vinyals,
  and Y.~W. Teh.
\newblock Attentive neural processes.
\newblock In {\em International Conference on Learning Representations}, 2018.

\bibitem{kim2022calibration}
S.~Kim and S.-Y. Yun.
\newblock Calibration of few-shot classification tasks: Mitigating
  misconfidence from distribution mismatch.
\newblock {\em IEEE Access}, 10:53894--53908, 2022.

\bibitem{kingma2014adam}
D.~P. Kingma and J.~Ba.
\newblock {Adam}: {A} method for stochastic optimization.
\newblock In {\em International Conference on Learning Representations}, 2015.

\bibitem{koller2009probabilistic}
D.~Koller and N.~Friedman.
\newblock {\em Probabilistic graphical models: principles and techniques}.
\newblock MIT press, 2009.

\bibitem{kuleshov2018accurate}
V.~Kuleshov, N.~Fenner, and S.~Ermon.
\newblock Accurate uncertainties for deep learning using calibrated regression.
\newblock In {\em International Conference on Machine Learning}, pages
  2796--2804, 2018.

\bibitem{kumar2018trainable}
A.~Kumar, S.~Sarawagi, and U.~Jain.
\newblock Trainable calibration measures for neural networks from kernel mean
  embeddings.
\newblock In {\em International Conference on Machine Learning}, pages
  2805--2814, 2018.

\bibitem{lakshminarayanan2017simple}
B.~Lakshminarayanan, A.~Pritzel, and C.~Blundell.
\newblock Simple and scalable predictive uncertainty estimation using deep
  ensembles.
\newblock {\em Advances in Neural Information Processing Systems}, 30, 2017.

\bibitem{maddox2019simple}
W.~J. Maddox, P.~Izmailov, T.~Garipov, D.~P. Vetrov, and A.~G. Wilson.
\newblock A simple baseline for {B}ayesian uncertainty in deep learning.
\newblock {\em Advances in Neural Information Processing Systems}, 32, 2019.

\bibitem{malik2019calibrated}
A.~Malik, V.~Kuleshov, J.~Song, D.~Nemer, H.~Seymour, and S.~Ermon.
\newblock Calibrated model-based deep reinforcement learning.
\newblock In {\em International Conference on Machine Learning}, pages
  4314--4323, 2019.

\bibitem{marx2022modular}
C.~Marx, S.~Zhao, W.~Neiswanger, and S.~Ermon.
\newblock Modular conformal calibration.
\newblock In {\em International Conference on Machine Learning}, pages
  15180--15195, 2022.

\bibitem{michelmore2020uncertainty}
R.~Michelmore, M.~Wicker, L.~Laurenti, L.~Cardelli, Y.~Gal, and M.~Kwiatkowska.
\newblock Uncertainty quantification with statistical guarantees in end-to-end
  autonomous driving control.
\newblock In {\em IEEE International Conference on Robotics and Automation},
  pages 7344--7350, 2020.

\bibitem{naeini2015obtaining}
M.~P. Naeini, G.~Cooper, and M.~Hauskrecht.
\newblock Obtaining well calibrated probabilities using {B}ayesian binning.
\newblock In {\em Twenty-Ninth AAAI Conference on Artificial Intelligence},
  2015.

\bibitem{nguyen2022transformer}
T.~Nguyen and A.~Grover.
\newblock Transformer neural processes: Uncertainty-aware meta learning via
  sequence modeling.
\newblock {\em arXiv preprint arXiv:2207.04179}, 2022.

\bibitem{niculescu2005predicting}
A.~Niculescu-Mizil and R.~Caruana.
\newblock Predicting good probabilities with supervised learning.
\newblock In {\em International Conference on Machine Learning}, pages
  625--632, 2005.

\bibitem{paszke2019pytorch}
A.~Paszke, S.~Gross, F.~Massa, A.~Lerer, J.~Bradbury, G.~Chanan, T.~Killeen,
  Z.~Lin, N.~Gimelshein, L.~Antiga, et~al.
\newblock Pytorch: An imperative style, high-performance deep learning library.
\newblock {\em Advances in Neural Information Processing Systems}, 32, 2019.

\bibitem{patacchiola2020bayesian}
M.~Patacchiola, J.~Turner, E.~J. Crowley, M.~O'Boyle, and A.~J. Storkey.
\newblock Bayesian meta-learning for the few-shot setting via deep kernels.
\newblock {\em Advances in Neural Information Processing Systems},
  33:16108--16118, 2020.

\bibitem{pearce2018high}
T.~Pearce, A.~Brintrup, M.~Zaki, and A.~Neely.
\newblock High-quality prediction intervals for deep learning: A
  distribution-free, ensembled approach.
\newblock In {\em International Conference on Machine Learning}, pages
  4075--4084, 2018.

\bibitem{platt1999probabilistic}
J.~Platt.
\newblock Probabilistic outputs for support vector machines and comparisons to
  regularized likelihood methods.
\newblock {\em Advances in Large Margin Classifiers}, 10(3):61--74, 1999.

\bibitem{rasmussen2005gaussian}
C.~E. Rasmussen and C.~K.~I. Williams.
\newblock {\em Gaussian Processes for Machine Learning}.
\newblock The MIT Press, 2005.

\bibitem{ravi2016optimization}
S.~Ravi and H.~Larochelle.
\newblock Optimization as a model for few-shot learning.
\newblock In {\em International Conference on Learning Representations}, 2017.

\bibitem{rothfuss2021meta}
J.~Rothfuss, D.~Heyn, A.~Krause, et~al.
\newblock Meta-learning reliable priors in the function space.
\newblock {\em Advances in Neural Information Processing Systems}, 34:280--293,
  2021.

\bibitem{sahoo2021reliable}
R.~Sahoo, S.~Zhao, A.~Chen, and S.~Ermon.
\newblock Reliable decisions with threshold calibration.
\newblock {\em Advances in Neural Information Processing Systems},
  34:1831--1844, 2021.

\bibitem{salem2020prediction}
T.~S. Salem, H.~Langseth, and H.~Ramampiaro.
\newblock Prediction intervals: Split normal mixture from quality-driven deep
  ensembles.
\newblock In {\em Conference on Uncertainty in Artificial Intelligence}, pages
  1179--1187, 2020.

\bibitem{schmidhuber:1987:srl}
J.~Schmidhuber.
\newblock Evolutionary principles in self-referential learning. on learning now
  to learn: The meta-meta-meta...-hook.
\newblock Master's thesis, Technische Universitat Munchen, Germany, 1987.

\bibitem{shahriari2015taking}
B.~Shahriari, K.~Swersky, Z.~Wang, R.~P. Adams, and N.~De~Freitas.
\newblock Taking the human out of the loop: A review of {B}ayesian
  optimization.
\newblock {\em Proceedings of the IEEE}, 104(1):148--175, 2015.

\bibitem{skafte2019reliable}
N.~Skafte, M.~J{\o}rgensen, and S.~Hauberg.
\newblock Reliable training and estimation of variance networks.
\newblock {\em Advances in Neural Information Processing Systems}, 32, 2019.

\bibitem{snell2017prototypical}
J.~Snell, K.~Swersky, and R.~Zemel.
\newblock Prototypical networks for few-shot learning.
\newblock {\em Advances in Neural Information Processing Systems}, 30, 2017.

\bibitem{snelson2005sparse}
E.~Snelson and Z.~Ghahramani.
\newblock Sparse {G}aussian processes using pseudo-inputs.
\newblock {\em Advances in Neural Information Processing Systems}, 18, 2005.

\bibitem{song2019distribution}
H.~Song, T.~Diethe, M.~Kull, and P.~Flach.
\newblock Distribution calibration for regression.
\newblock In {\em International Conference on Machine Learning}, pages
  5897--5906, 2019.

\bibitem{tagasovska2019single}
N.~Tagasovska and D.~Lopez-Paz.
\newblock Single-model uncertainties for deep learning.
\newblock {\em Advances in Neural Information Processing Systems}, 32, 2019.

\bibitem{tan2014time}
S.~C. Tan and J.~P.~S. Lau.
\newblock Time series clustering: A superior alternative for market basket
  analysis.
\newblock In {\em Proceedings of International Conference on Advanced Data and
  Information Engineering}, pages 241--248. Springer, 2014.

\bibitem{tossou2019adaptive}
P.~Tossou, B.~Dura, F.~Laviolette, M.~Marchand, and A.~Lacoste.
\newblock Adaptive deep kernel learning.
\newblock {\em arXiv preprint arXiv:1905.12131}, 2019.

\bibitem{tran2019calibrating}
G.-L. Tran, E.~V. Bonilla, J.~Cunningham, P.~Michiardi, and M.~Filippone.
\newblock Calibrating deep convolutional {G}aussian processes.
\newblock In {\em The 22nd International Conference on Artificial Intelligence
  and Statistics}, pages 1554--1563, 2019.

\bibitem{vinyals2016matching}
O.~Vinyals, C.~Blundell, T.~Lillicrap, D.~Wierstra, et~al.
\newblock Matching networks for one shot learning.
\newblock In {\em Advances in Neural Information Processing Systems}, pages
  3630--3638, 2016.

\bibitem{vovk2020conformal}
V.~Vovk, I.~Petej, P.~Toccaceli, A.~Gammerman, E.~Ahlberg, and L.~Carlsson.
\newblock Conformal calibrators.
\newblock In {\em Conformal and Probabilistic Prediction and Applications},
  pages 84--99, 2020.

\bibitem{wilson2016deep}
A.~G. Wilson, Z.~Hu, R.~Salakhutdinov, and E.~P. Xing.
\newblock Deep kernel learning.
\newblock In {\em International Conference on Artificial Intelligence and
  Statistics}, pages 370--378, 2016.

\bibitem{yang2022calibrating}
P.~Yang, S.~Ren, Y.~Zhao, and P.~Li.
\newblock Calibrating cnns for few-shot meta learning.
\newblock In {\em Proceedings of the IEEE/CVF Winter Conference on Applications
  of Computer Vision}, pages 2090--2099, 2022.

\bibitem{yu2020mopo}
T.~Yu, G.~Thomas, L.~Yu, S.~Ermon, J.~Y. Zou, S.~Levine, C.~Finn, and T.~Ma.
\newblock {MOPO}: Model-based offline policy optimization.
\newblock {\em Advances in Neural Information Processing Systems},
  33:14129--14142, 2020.

\bibitem{zhao2020individual}
S.~Zhao, T.~Ma, and S.~Ermon.
\newblock Individual calibration with randomized forecasting.
\newblock In {\em International Conference on Machine Learning}, pages
  11387--11397, 2020.

\end{thebibliography}

\end{document}